\documentclass{article}

\usepackage{microtype}
\usepackage{graphicx}
\usepackage{subcaption}
\usepackage{booktabs} 

\usepackage{hyperref}

\usepackage[accepted]{icml2026}

\usepackage{amsmath}
\usepackage{amssymb}
\usepackage{mathtools}
\usepackage{amsthm}

\usepackage[capitalize,noabbrev]{cleveref}

\theoremstyle{plain}

\theoremstyle{definition}

\theoremstyle{remark}

\usepackage[textsize=tiny]{todonotes}

\usepackage{amsmath}    
\usepackage{amssymb}    
\usepackage{amsfonts}   

\usepackage{booktabs}   
\usepackage{siunitx}    
\usepackage{makecell}   
\usepackage{adjustbox}  
\usepackage{placeins}

\usepackage{tikz}
\usetikzlibrary{arrows.meta, positioning}

\icmltitlerunning{Functional Distribution Networks (FDN)}

\begin{document}

\twocolumn[
  \icmltitle{Functional Distribution Networks (FDN)}



  \icmlsetsymbol{equal}{*}

  \begin{icmlauthorlist}
    \icmlauthor{Omer Haq}{equal,yyy}
  \end{icmlauthorlist}

  \icmlaffiliation{yyy}{Independent Researcher, Baltimore, MD, USA}

  \icmlcorrespondingauthor{Omer Haq}{haqomer1@gmail.com}

  \icmlkeywords{Machine Learning, ICML}

  \vskip 0.3in
]



\printAffiliationsAndNotice{}  

\begin{abstract}
Modern probabilistic regressors often remain overconfident under distribution shift. Functional Distribution Networks (FDN) place input-conditioned distributions over network weights, producing predictive mixtures whose dispersion adapts to the input; we train them with a Monte Carlo $\beta$--ELBO objective. We pair FDN with an evaluation protocol that separates interpolation from extrapolation and emphasizes simple OOD sanity checks. On controlled 1D tasks and small/medium UCI-style regression benchmarks, FDN remains competitive in accuracy with strong Bayesian, ensemble, dropout, and hypernetwork baselines, while providing strongly input-dependent, shift-aware uncertainty and competitive calibration under matched parameter and update budgets.
\end{abstract}

\section{Introduction}
Modern neural predictors are often deployed under dataset shift, where
test inputs depart from the training distribution. In such regimes,
deterministic networks and common stochastic heuristics frequently become
\emph{overconfident}, assigning high probability to incorrect predictions
far from the training support and undermining reliable decision making.
Bayesian Neural Networks (BNNs), MLPs with dropout, Deep Ensembles, and
Hypernetworks are strong practical baselines, yet they can still
under-react under shift or require substantial ensembling or sampling to
achieve robust uncertainty behavior
\citep{quinonero2009dataset,ovadia2019can,guo2017calibration,
mackay1992practical,neal1996bayesian,graves2011practical,
blundell2015weight,gal2016dropout,lakshminarayanan2017simple,
ha2017hypernetworks}.

This motivates architectures that are explicitly \emph{uncertainty-aware}
and \emph{calibrated}, producing sharp predictions in-distribution (ID)
while widening appropriately under distribution shift. We pursue this
goal with \emph{Functional Distribution Networks (FDN)}, which place
input-conditioned distributions over network weights, allowing
predictive uncertainty to adapt locally in input space. FDN amortizes an
input-conditional posterior $q_\phi(\theta\mid x)$ via small
hypernetworks and is trained using a Monte Carlo likelihood and a
$\beta$-ELBO objective.

Empirically, under matched parameter and inference budgets, FDN is
ID-competitive and exhibits strong calibration under shift. On
controlled 1D regression tasks, predictive variance closely tracks
error growth in- and out-of-distribution, while on more challenging
oscillatory shifts FDN preserves rank calibration and increases
uncertainty OOD, albeit with some under-scaling. We further evaluate FDN
on standard UCI-style regression benchmarks—Airfoil Self-Noise, Combined
Cycle Power Plant, and Energy Efficiency
\citep{uci_airfoil,uci_ccpp,uci_energy}—where it typically widens
uncertainty under feature-based shifts while maintaining competitive
accuracy and calibration relative to strong Bayesian, ensemble, and
hypernetwork baselines.

\paragraph{Overview.}
Rather than treating weights as fixed (or globally random), FDN places an input-conditioned distribution over weights:
\[
\theta \mid x \sim p(\theta \mid x), \qquad y \mid x, \theta \sim p(y \mid x, \theta).
\]
For tractability we fix an input-agnostic prior $p(\theta \mid x) = p_0(\theta) = \mathcal{N}(0, \sigma_0^2 I)$, so all input dependence is carried by an amortized posterior $q_\phi(\theta \mid x)$ implemented via small Hypernetworks. Many standard uncertainty-aware architectures can be cast in this template via different choices of $q_{\phi}(\theta \mid x)$; we detail these instantiations in Appendix~\ref{app:q_phi}. Sampling $\theta \sim q_\phi(\theta \mid x)$ yields locally adaptive functions, and as $x$ moves off the training support the induced weight distribution can broaden, producing wider and more appropriately uncertain predictive densities.

\paragraph{Training and evaluation.}
We train with a Monte Carlo $\beta$--ELBO over $q_\phi(\theta \mid x)$ and also report an IWAE variant; explicit expressions for the layer-wise KL and its LP-FDN factorization are given in the Appendix~\ref{app:training}. Our evaluation protocol splits test points into interpolation (ID) and extrapolation (OOD) regions and summarizes shift via deltas $\Delta(\cdot) = \mathbb{E}_{\text{OOD}}[\cdot] - \mathbb{E}_{\text{ID}}[\cdot]$, focusing on $\Delta\mathrm{MSE}$, $\Delta\mathrm{Var}$, and $\Delta\mathrm{CRPS}$ together with MSE--variance fits and risk--coverage curves.

\noindent\textbf{Contributions.}
(i) \textbf{Model.} We introduce Functional Distribution Networks (FDN), a simple module that amortizes input-conditioned weight distributions via small Hypernetworks, in two variants: IC-FDNet (conditioning on $x$) and LP-FDNet (conditioning layer-wise on previous activations). (ii) \textbf{Protocol.} We propose a small-suite extrapolation protocol that targets $\Delta\mathrm{Var} > 0$ under shift and complements it with calibration diagnostics (MSE--variance slope, rank correlation, and risk--coverage). (iii) \textbf{Empirical study.} Under matched parameter, update, and predictive-sample budgets, we benchmark FDN against strong Bayesian, ensemble, dropout, and hypernetwork baselines on controlled 1D function families and UCI-style regression tasks, showing that FDN remains ID-competitive while providing practically useful, shift-aware uncertainty.

\noindent\textbf{Scope.}
We focus on low-dimensional regression with homoscedastic scalar Gaussian heads and relatively shallow backbones in order to isolate ID vs.\ OOD behavior under tightly matched budgets. All code, configurations, and scripts to reproduce the experiments will be released upon acceptance.

\section{Related Work} \label{sec:related}
\textbf{Uncertainty in neural regression.} BNN place distributions over weights and infer posteriors via variational approximations or MCMC \citep{mackay1992practical, neal1996bayesian, graves2011practical, blundell2015weight}. Deep Ensembles average predictions from independently trained networks and are a strong practical baseline \citep{lakshminarayanan2017simple, maddox2019simple}. MLP (with dropout) interprets dropout at test time as approximate Bayesian inference \citep{gal2016dropout}. Heteroscedastic regression learns input-dependent output variance but retains deterministic weights \citep{nix1994estimating, kendall2017uncertainties}.

\textbf{Hypernetworks and conditional weight generation.}
Hypernetworks generate the weights of a primary network using an auxiliary network \citep{ha2017hypernetworks}; related work explores dynamic, input-conditioned filters and conditional computation \citep{debrabandere2016dynamic,brock2018smash}. Bayesian/uncertainty-aware Hypernetworks place distributions over generated weights and train them variationally~\citep{krueger2017bayesian}. Our FDN differs by explicitly conditioning the weight distributions on the current input (or intermediate activations) in order to modulate epistemic uncertainty itself, not only deterministic weights. This input-aware weight stochasticity is what enables FDN to widen uncertainty under distributional shift while maintaining competitive ID accuracy under matched budgets.


\textbf{Calibration and OOD behavior.}
Proper scoring rules such as the continuous ranked probability score (CRPS) are strictly proper and reward calibrated predictive distributions \citep{gneiting2007strictly}. Empirical studies highlight overconfidence under dataset shift and introduce OOD benchmarks \citep{ovadia2019can}. Our evaluation therefore separates interpolation from extrapolation and uses the monotonic relationship between per-sample squared error (MSE) and predicted variance as a simple diagnostic calibration check.

\paragraph{Positioning.} Compared to BNNs, FDN sidesteps global posteriors by amortizing \emph{local} weight distributions $q_\phi(\theta\mid x)$. Compared to Deep Ensembles, FDN uses shared parameters and stochastic generation instead of replicating full models. Compared to Hypernetworks, FDN explicitly models \emph{uncertainty over} generated weights and regularizes it with a KL prior, enabling principled OOD expansion.

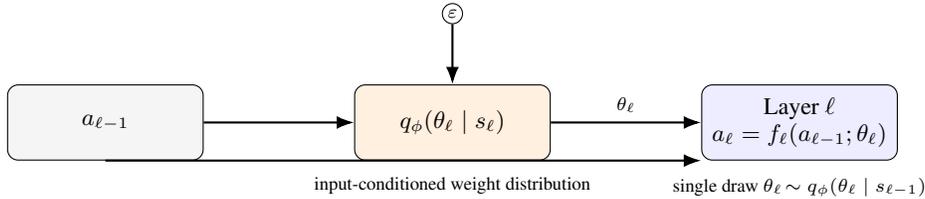
\begin{figure*}[t]
\centering
\begin{tikzpicture}[
  node distance=1.3cm and 2.0cm,
  box/.style={draw, rounded corners, minimum width=2.6cm, minimum height=1.0cm, align=center, font=\small},
  main/.style={box, fill=blue!7},
  hyper/.style={box, fill=orange!12},
  input/.style={box, fill=gray!8},
  arrow/.style={-Latex, thick},
  lbl/.style={font=\scriptsize}
]

\node[input] (aprev) {$a_{\ell-1}$};  
\node[hyper, right=of aprev] (hyper) {$q_\phi(\theta_\ell \mid s_{\ell})$};
\node[main,  right=of hyper] (layer) {Layer $\ell$\\$a_\ell = f_\ell(a_{\ell-1};\theta_\ell)$};

\node[draw, circle, inner sep=1pt, above=0.8cm of hyper] (eps) {\scriptsize$\varepsilon$};

\draw[arrow] (aprev) -- (hyper);
\draw[arrow] (aprev) |- (layer.south west);
\draw[arrow] (eps) -- (hyper.north);
\draw[arrow] (hyper) -- node[midway, above, lbl] {$\theta_\ell$} (layer);

\node[lbl, below=0.1cm of hyper] {input-conditioned weight distribution};
\node[lbl, below=0.1cm of layer]  {single draw $\theta_\ell\!\sim q_\phi(\theta_\ell \mid s_{\ell-1})$};

\end{tikzpicture}
\caption{Single-layer view of FDNet. For layer $\ell$, the previous
activation $a_{\ell-1}$ (with $a_0 = x$ for the first layer) is fed
to both the hypernetwork and the main layer. For IC-FDNet $s_{\ell}=x$ while for LP-FDNet $s_{\ell}=a_{\ell-1}$. The hypernetwork takes
$s_{\ell}$ and a random draw $\varepsilon$ to generate weights/ biases (Gaussian Head with reparametrization technique) $\theta_\ell=(W_\ell, b_\ell)$, which are then used by the main layer
$a_\ell = f_\ell(a_{\ell-1};\theta_\ell)$.}
\label{fig:ic-fdnet-layer}
\end{figure*}

\section{Method} 
For convenience, a summary of the main symbols and dimensions used throughout the paper is provided in Table~\ref{tab:notation-core} in Appendix~\ref{app:notation}.

\subsection{Preliminaries}
We model $y\in\mathbb{R}^{d_y}$ with a neural network
$f_\theta:\mathbb{R}^{d_x}\!\to\!\mathbb{R}^{D_y}$ whose final layer
parametrizes a Gaussian predictive head $(\mu_\theta(x),\Sigma_\theta(x))$,
yielding
\[
p(y\mid x,\theta)=\mathcal{N}\!\big(y;\mu_\theta(x),\Sigma_\theta(x)\big).
\]

In the main experiments we use a
homoscedastic Gaussian likelihood
$p(y\mid x,\theta)=\mathcal{N}\!\bigl(y;\mu_\theta(x),\sigma^2\bigr)$
with fixed variance $\sigma^2$, shared across all models.
This removes aleatoric uncertainty and isolates epistemic uncertainty
arising from stochastic weights.
Under this likelihood, the $\beta$--ELBO reduces (up to constants) to a
weighted squared-error term plus a KL penalty, yielding a
$MSE+\beta\,KL$ objective.
We use the same Gaussian head for all baselines to ensure a fair comparison.

More general likelihoods are discussed in Appendix~\ref{app:hetero}.

\subsection{FDN: Input-Conditioned Weight Distributions}
We drop explicit context and condition only on signals from the network itself. 
For each layer $\ell$, FDN places a diagonal-Gaussian over its weights whose parameters are produced by a small Hypernetwork $A_\ell(\cdot)$. 
We choose the conditioning signal
\[
s_{\ell}^{(k)} \in 
\begin{cases}
x, & \text{IC-FDN}\\
a_{\ell-1}^{(k)}, & \text{LP-FDN (with } a_{0}^{(k)}=x \text{)},
\end{cases}
\]
and set
\[
\begin{aligned}
(\mu_{W,\ell},\,\rho_{W,\ell},\,\mu_{b,\ell},\,\rho_{b,\ell})
&= A_\ell\!\left(s_{\ell}^{(k)}\right),
\end{aligned}
\]
where $\sigma_{.,\ell} = \varepsilon + \operatorname{softplus}(\rho_{.,\ell})$. with a small variance floor $\varepsilon=10^{-3}$ for numerical stability. 
Sampling then proceeds via the standard Gaussian reparameterization trick. The conditioning signal is chosen as $s_\ell = x$ for IC-FDNet and
$s_\ell = a_{\ell-1}^{(k)}$ for LP-FDNet, where in both cases the layer
output is
$a_\ell^{(k)} = f_\ell(a_{\ell-1}^{(k)}; W_\ell^{(k)}, b_\ell^{(k)})$
(sequential across layers). Figure~\ref{fig:ic-fdnet-layer} illustrates
a single FDN layer: a small hypernetwork maps $s_\ell$ to the parameters
of a Gaussian distribution over weights and biases; a sample
$\theta_\ell = (W_\ell, b_\ell)$ from this distribution is then used to
compute $a_\ell = f_\ell(a_{\ell-1}; \theta_\ell)$.

\paragraph{Variational family (compact).}
FDN uses a layer-wise diagonal-Gaussian over weights, conditioned on $s_\ell\!\in\!\{x,\ a_{\ell-1}^{(k)}\}$:
\[
q_\phi(\theta \mid x)
= \prod_{\ell=1}^{L}
\mathcal{N}\!\big(
  (W_\ell,b_\ell);
  (\mu_{W,\ell},\mu_{b,\ell}),
  \operatorname{diag}(\sigma_{W,\ell}^2,\sigma_{b,\ell}^2)
\big).
\]
In compact form (concatenating all layers),
\[
q_\phi(\theta\mid x)=\mathcal{N}\!\big(\theta;\,\mu_\phi(x),\,\operatorname{diag}\sigma_\phi^2(x)\big),
\]
and the predictive density is the Monte Carlo mixture
\[
p(y\mid x)\;\approx\;\frac{1}{K}\sum_{k=1}^{K} p\!\big(y\mid x,\theta^{(k)}\big),\qquad \theta^{(k)}\sim q_\phi(\theta\mid x).
\]

\paragraph{Prior and regularization.}
We regularize $q_\phi(\theta \,|\, x)$ toward a simple reference
$p_0(\theta)=\prod_{\ell}\mathcal{N}(0,\sigma_0^2 I)$ via a $\beta$-weighted
KL term (we use $\sigma_0=1$ in all experiments).
For diagonal Gaussians,
\[
KL\!\left(
\mathcal{N}(\mu,\operatorname{diag}\sigma^2)
\,\middle\|\,
\mathcal{N}(0,\sigma_0^2 I)
\right)
= \tfrac{1}{2}\sum_j KL_j,
\]
where
$KL_j = (\sigma_j^2+\mu_j^2)/\sigma_0^2 - 1 - \log(\sigma_j^2/\sigma_0^2)$.
We sum this over all layers and parameters (both $W_\ell$ and $b_\ell$).

Algorithm~\ref{alg:fdn-unified} (Appendix~\ref{app:alg}) summarizes training ($\beta$-ELBO with re-parameterized gradients) and inference (Monte-Carlo mixtures over weight draws) for both IC-FDN and LP-FDN.

\paragraph{IC-FDNet vs.\ LP-FDNet.}
We study two conditioning schemes. \emph{IC-FDNet} (Input-Conditioned)
uses the raw input $x$ at every layer, $s_\ell(x) = x$, so all layers
see the same features when sampling weights. \emph{LP-FDNet}
(Layer-Progressive) instead uses hidden activations, with $s_0(x) = x$
and $s_\ell(x) = a_{\ell-1}(x)$ for $\ell \ge 1$, yielding a depth-aware
conditioning scheme. Empirically, the two variants achieve similar
in-distribution accuracy across our benchmarks, while LP-FDNet often
produces somewhat larger increases in predictive variance under
distribution shift (larger $\Delta\text{Var}$) at comparable MSE.

\begin{figure*}
  \centering
  \captionsetup[subfigure]{justification=centering,singlelinecheck=false,skip=2pt}

  \begin{subfigure}[t]{0.98\linewidth}
    \centering
    \includegraphics[width=0.95\linewidth]{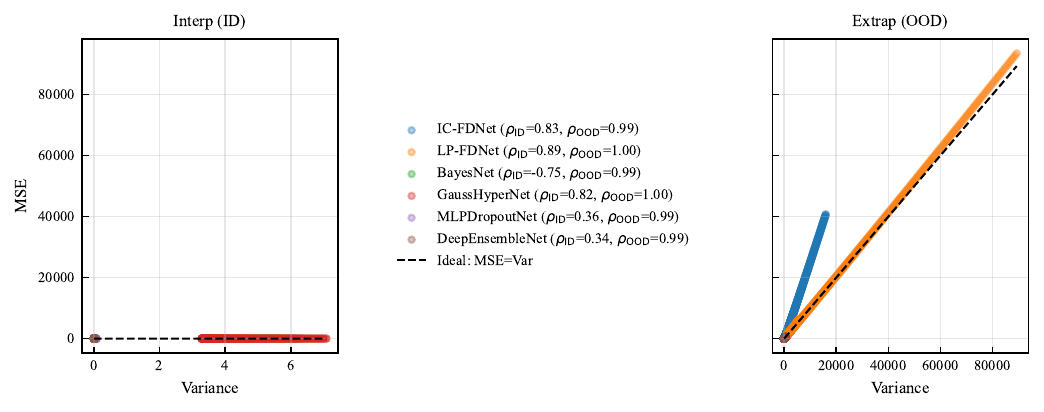}
    \subcaption{Step: $H(x)$}
    \label{fig:toy-scatter-step}
  \end{subfigure}

  \vspace{0.4em}

  \begin{subfigure}[t]{0.98\linewidth}
    \centering
    \includegraphics[width=0.95\linewidth]{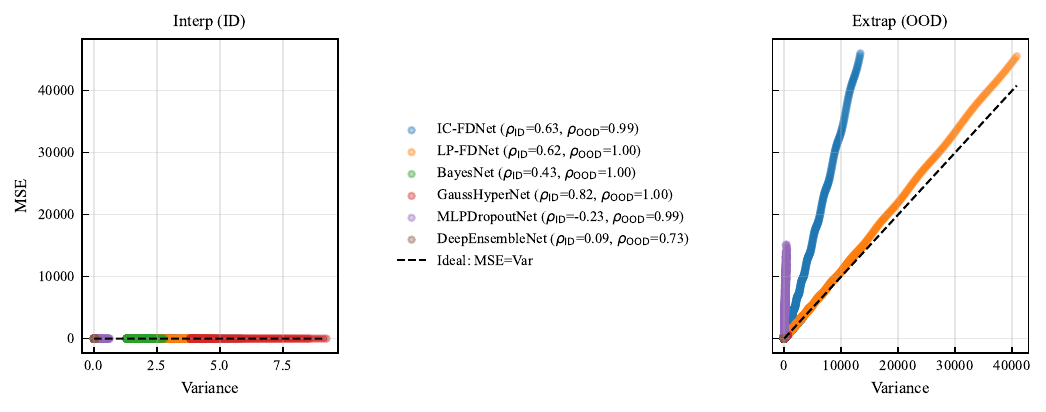}
    \subcaption{Sine: $1.54\,\sin(2.39\,x)$}
    \label{fig:toy-scatter-sine}
  \end{subfigure}

  \vspace{0.4em}

  \begin{subfigure}[t]{0.98\linewidth}
    \centering
    \includegraphics[width=0.95\linewidth]{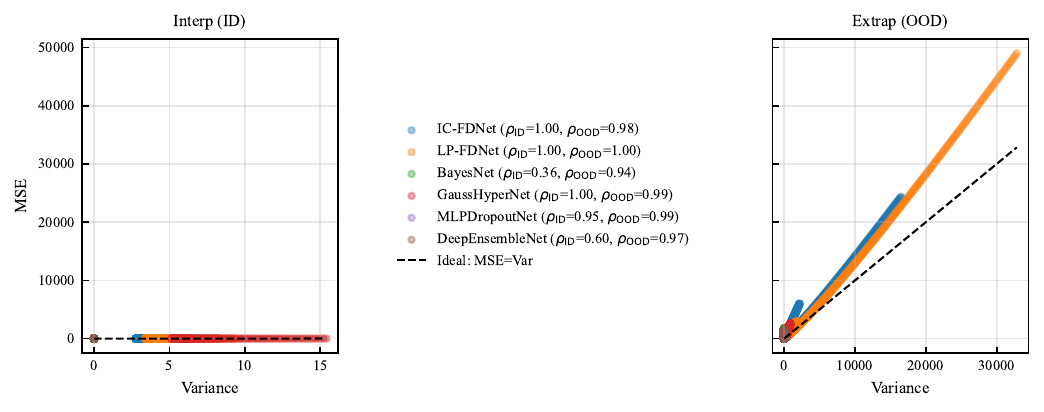}
    \subcaption{Quadratic: $0.43x^{2}-0.41$}
    \label{fig:toy-scatter-quad}
  \end{subfigure}

    \caption{
    MSE vs.\ predicted variance on three 1D regression tasks
    (rows: step, sine, quadratic).
    Left/right panels in each row show interpolation (ID) and extrapolation
    (OOD) test points with shared axes; the dashed line marks the ideal
    $\mathrm{MSE}=\mathrm{Var}$. Legends report per-model Spearman's
    $\rho$ in each region.
    }
    
  \label{fig:toy-msevar-grid}
\end{figure*}

\begin{figure*}
  \centering
  \captionsetup[subfigure]{justification=centering,singlelinecheck=false,skip=2pt}

  \begin{subfigure}[t]{0.98\linewidth}
    \centering
    \includegraphics[width=0.95\linewidth]{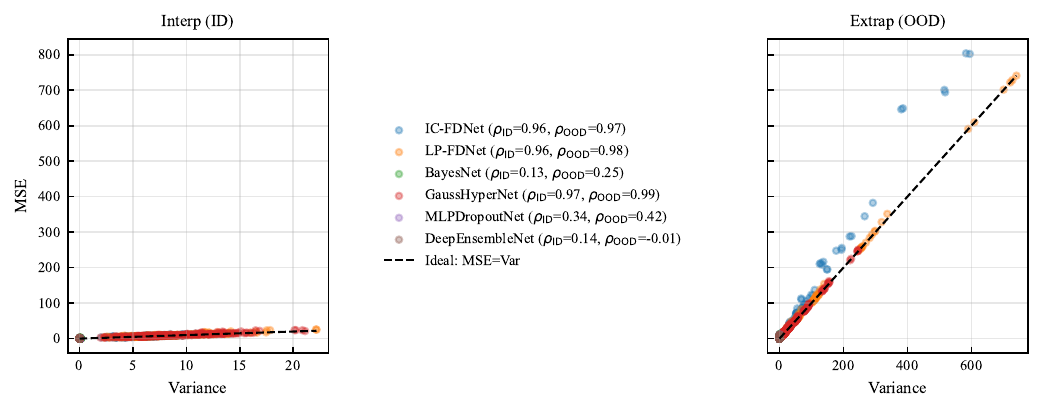}
    \subcaption{Airfoil Self-Noise}
    \label{fig:real-scatter-airfoil}
  \end{subfigure}

  \vspace{0.4em}

  \begin{subfigure}[t]{0.98\linewidth}
    \centering
    \includegraphics[width=0.95\linewidth]{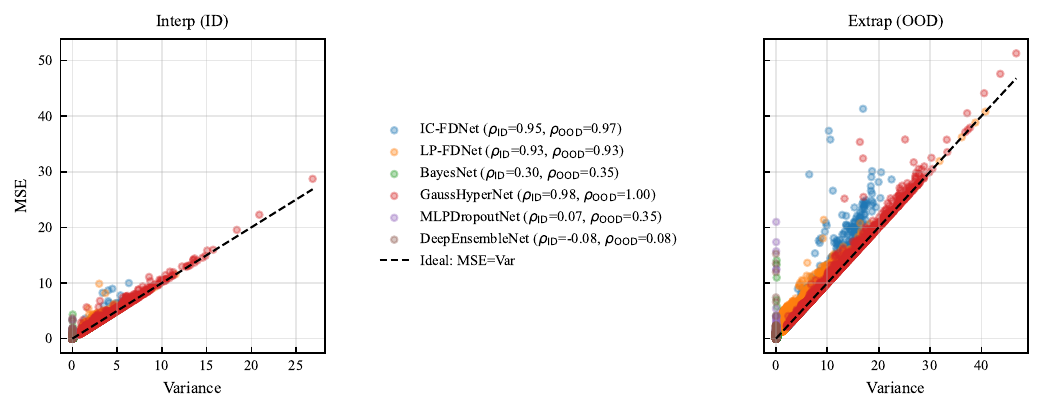}
    \subcaption{CCPP Power Plant}
    \label{fig:real-scatter-ccpp}
  \end{subfigure}

  \vspace{0.4em}

  \begin{subfigure}[t]{0.98\linewidth}
    \centering
    \includegraphics[width=0.95\linewidth]{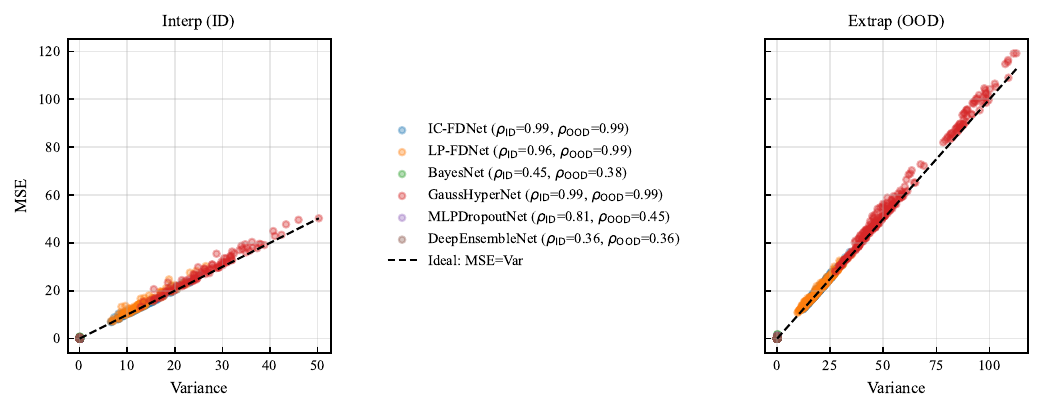}
    \subcaption{Energy Efficiency Heating}
    \label{fig:real-scatter-energy}
  \end{subfigure}

    \caption{
    MSE vs.\ predicted variance on three real regression datasets:
    Airfoil Self-Noise, CCPP Power Plant, and Energy Efficiency (heating).
    For each dataset, left/right panels show ID and OOD test points with
    shared axes, and the dashed line marks $\mathrm{MSE}=\mathrm{Var}$.
    Legends report per-model Spearman's $\rho$.
    }

  \label{fig:real-msevar-grid}
\end{figure*}

\begin{figure*}
  \centering

  \begin{subfigure}[t]{0.72\linewidth}
    \centering
    \includegraphics[width=\linewidth]{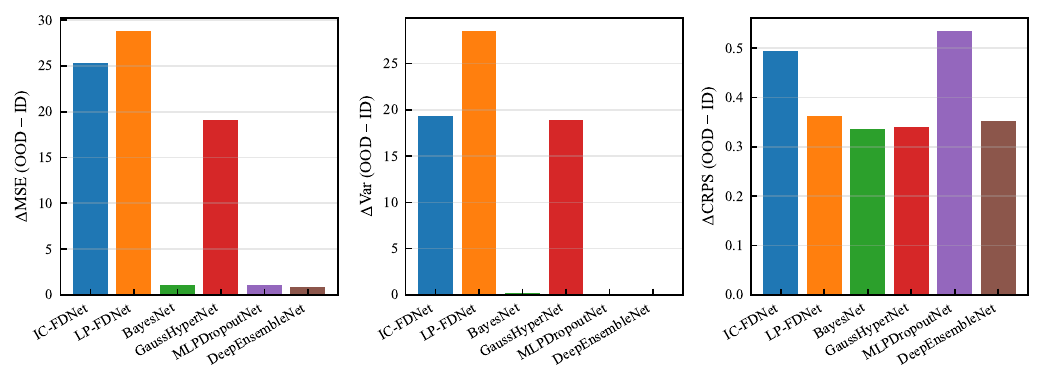}
    \subcaption{Airfoil Self-Noise}
    \label{fig:bars-crps-airfoil}
  \end{subfigure}

  \vspace{0.35em}

  \begin{subfigure}[t]{0.72\linewidth}
    \centering
    \includegraphics[width=\linewidth]{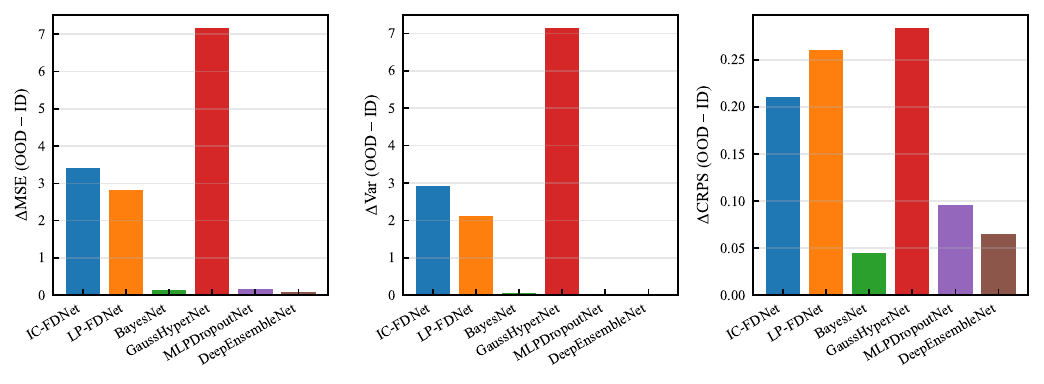}
    \subcaption{CCPP Power Plant}
    \label{fig:bars-crps-ccpp}
  \end{subfigure}

  \vspace{0.35em}

  \begin{subfigure}[t]{0.72\linewidth}
    \centering
    \includegraphics[width=\linewidth]{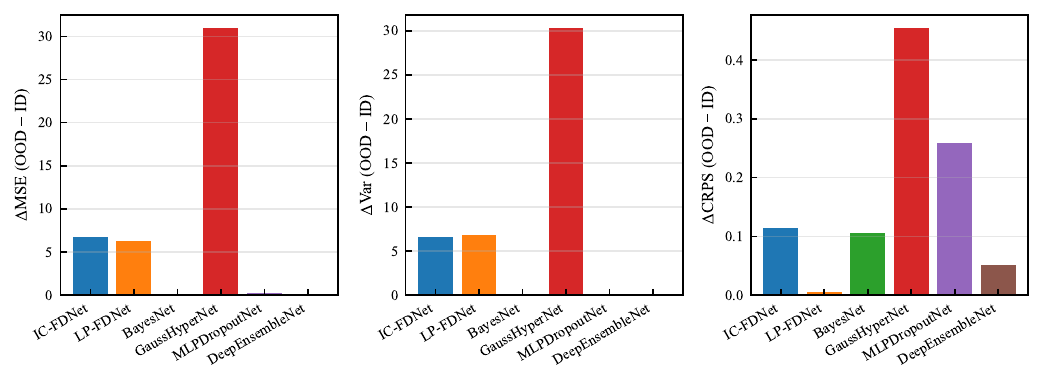}
    \subcaption{Energy Efficiency Heating}
    \label{fig:bars-crps-quad}
  \end{subfigure}
  
    \caption{
    ID$\!\to$OOD deltas on the real regression datasets.
    Bars show $\Delta\mathrm{MSE}$,
    $\Delta\mathrm{Var}$, and $\Delta\mathrm{CRPS}$ for Airfoil, CCPP,
    and Energy. Well-behaved models exhibit large positive
    $\Delta\mathrm{Var}$ together with moderate $\Delta\mathrm{MSE}$ and
    $\Delta\mathrm{CRPS}$, indicating that uncertainty widens under shift
    while accuracy and calibration degrade gracefully.
    }
  \label{fig:bars-crps-vertical}
\end{figure*}

\section{Experiments} 

\paragraph{Tasks and splits.}
We evaluate FDN on (i) controlled 1D regression tasks where the
ground-truth function is known and interpolation vs.\ extrapolation
is precisely defined in input space, and (ii) standard UCI-style regression benchmarks with feature-based ID/OOD splits.
In both settings we report metrics separately on the interpolation
(ID) region, the extrapolation (OOD) region, and the aggregated test
set, and we summarize distribution shift by deltas
$\Delta(\cdot)=\mathbb{E}_{\mathrm{OOD}}[\cdot]-
\mathbb{E}_{\mathrm{ID}}[\cdot]$.

\subsection{Toy function families and ID/OOD protocol}
\label{sec:toy-data}

We first benchmark on 1D toy regression tasks, where the ground-truth
function $f : \mathbb{R} \to \mathbb{R}$ is known. For each run we
select one of three families:
\begin{itemize}
\item \textbf{Step}: $f(x) = H(x) = \mathbf{1}\{x \ge x_0\}$, with a discontinuity at $x_0 = 0$.
\item \textbf{Sine}: $f(x) = A \sin(\omega x)$, with amplitude $A$ and frequency $\omega$.
\item \textbf{Quadratic}: $f(x) = ax^2 + b$, with $(a,b)$ fixed across runs.
\end{itemize}
Unless otherwise stated we use the specific instantiations shown in
Figure~\ref{fig:toy-msevar-grid} (step $H(x)$, sine
$1.54 \sin(2.39x)$, and quadratic $0.43x^2 - 0.41$) to match all
reported plots. We define a symmetric interpolation region
$R_{\mathrm{interp}} = [-\ell,\ell]$ and an extrapolation region
$R_{\mathrm{extrap}} = \mathbb{R} \setminus R_{\mathrm{interp}}$.
Training and validation inputs are sampled uniformly from
$R_{\mathrm{interp}}$; test inputs cover both $R_{\mathrm{interp}}$
and $R_{\mathrm{extrap}}$ on a dense grid. We report all metrics
separately on the ID split (test points in $R_{\mathrm{interp}}$),
the OOD split (test points outside $R_{\mathrm{interp}}$),
and on the full test set. This protocol ensures that ``OOD''
strictly corresponds to extrapolation in input space.

\subsection{Real regression datasets and ID/OOD splits}
\label{sec:real-data}
To test FDNs beyond 1D toy functions we use three standard UCI regression datasets:
Airfoil Self-Noise, Combined Cycle Power Plant (CCPP), and Energy Efficiency (heating load as the primary target). For each dataset we follow a consistent preprocessing and ID/OOD protocol: we choose a single ``ID feature'' $z$ with a natural interpretation (e.g., frequency for Airfoil, ambient temperature for CCPP, relative compactness for Energy), define the interpolation band as the 20th--80th percentiles of this feature and treat the extremes as extrapolation, split train/validation/test by quantiles of $z$, and standardize all inputs and targets using training statistics. We fix a single train/validation/test split shared across all methods and seeds. Exact dataset sizes are summarized in Table~\ref{tab:data_splits} (Appendix~\ref{app:baseline}).

\subsection{Complexity, capacity, and fairness}

\paragraph{Baselines.}
We evaluate four stochastic baselines: MLP with dropout
(\textbf{MLPDropoutNet}), Deep Ensemble of MLP
(\textbf{DeepEnsemblesNet}), Variational BNN (\textbf{BayesNet}), and
Gaussian Hypernetwork (\textbf{GaussianHyperNet}). Because our study
centers on calibrated \emph{predictive distributions} (CRPS, MSE–Var
slope/intercept, AURC), we omit the \textbf{Deterministic MLP} and
the (input-conditioned) \textbf{Hypernetwork} from the main
uncertainty analysis; their ID/OOD MSE is comparable to
Ensembles/Dropout. Training details (optimizer, batch size, learning
rate, prior scale $\sigma_0$, variance floor $\sigma_{\min}$) appear
in Table~\ref{tab:mc_k} (Appendix~\ref{app:baseline}).

\noindent\textbf{Link-budget.}
We consider networks with a single hidden layer and fix the parameter
budget to \(P \approx 1000\) (±5\%) for all models, counting
\emph{all} trainable parameters, including any Hypernetwork
components; counts appear in Table~\ref{tab:model_grid}
(Appendix~\ref{app:baseline}). To equalize the \emph{update} budget,
ensembles with \(M\) members use \emph{epoch-split} training (epochs
divided by \(M\)). For non-ensemble networks we use one Monte Carlo
draw per update (\(K=1\)), keeping per-step cost comparable and the
total number of parameter updates matched across models.

\textit{Note:} To hit the $P$ target, MLPDropoutNet uses widened
hidden layers, increasing capacity/expressive power and potentially
improving ID MSE independent of uncertainty quality; hence our
emphasis on calibration-centric metrics.

\paragraph{Computational complexity and scalability.}
For all models, the per-epoch training cost scales linearly in the
number of examples $N$ and quadratically in the layer widths,
$O\!\left(N \sum_{\ell} d_{\ell-1} d_{\ell}\right)$. Both IC-FDN and
LP-FDN inherit this scaling from the base MLP: the per-layer
Hypernetworks add only a small constant-factor overhead because their
hidden width $h_{\mathrm{hyp}}$ and the Monte Carlo count $K$ are
fixed (parameter count remains $O(d_{\ell-1} d_\ell)$ per layer),
matching the baselines. A full discussion of parameter-count and architectural trade-offs for larger
models are given in Appendix~\ref{app:complexity}.

\subsection{Metrics and calibration diagnostics}
\label{sec:metrics}

We assess models with standard metrics for accuracy, uncertainty, and
calibration: mean squared error (MSE), predictive variance, continuous ranked probability score
(CRPS), risk–coverage curves (area under the risk–coverage curve,
AURC), and ID$\!\to$OOD deltas $(\Delta\text{MSE},
\Delta\text{Var}, \Delta\text{CRPS})$. For calibration we use both
rank- and scale-based diagnostics: (i) Spearman correlation
$\rho(\mathrm{Var},\mathrm{MSE})$ between per-point variance and
squared error, and (ii) a linear fit
$\mathrm{MSE} \approx a + b\,\mathrm{Var}$, with the ideal
calibration corresponding to $a \approx 0$ and $b \approx 1$. We
visualize these diagnostics via MSE–Var scatter plots in the ID and
OOD regions for toy tasks (Figure~\ref{fig:toy-msevar-grid}) and real
datasets (Figure~\ref{fig:real-msevar-grid}), together with grouped
bar charts of ID$\!\to$OOD deltas on the real tasks
(Figure~\ref{fig:bars-crps-vertical}). Precise definitions and Monte
Carlo estimators for all metrics are collected in
Appendix~\ref{app:metrics-calib}.


\paragraph{Representative seed selection.}
To avoid cherry-picking, all qualitative plots are based on a
\emph{representative} seed selected by a fixed aggregation rule. For
each dataset and configuration, we run all models over multiple random
seeds (20 for toy tasks, 100 for Airfoil, and 3 for the remaining UCI
datasets) and compute per-seed summary metrics (e.g., ID/OOD MSE,
variance, $\Delta\mathrm{CRPS}$, AURC). For IC-FDNet (or LP-FDNet), we
take the coordinate-wise median of these metrics and select the seed
whose metric vector is closest (Euclidean distance) to this median.
All qualitative plots use this representative seed under the same rule
across datasets, ensuring the shown behavior is typical rather than
hand-picked. For Airfoil, Figure~\ref{fig:seed-agg-mse-ood-var-ood}
(Appendix~\ref{app:baseline}) confirms that the selected seed lies close
to the overall across-seed trend.

\subsection{Results}

\paragraph{Toy tasks.}
Across the three toy tasks (Tables~\ref{tab:step-unified}--\ref{tab:quad-unified};
Figure~\ref{fig:toy-msevar-grid}), FDN’s core strength is scale calibration
under smooth shifts. On the step and quadratic tasks, IC-/LP-FDNet achieve
MSE--Var slopes closer to the ideal $b \approx 1$ with strong rank agreement
(Spearman $\rho$ close to 1) and large positive $\Delta\text{Var}$, so
predictive variance increases in lock-step with difficulty. This comes at
the cost of higher AURC and $\Delta\text{CRPS}$ than the sharpest baselines
on the step task, while on the quadratic task their AURC and
$\Delta\text{CRPS}$ are broadly comparable. In contrast, several classical
baselines that fit ID sharply (e.g., Deep Ensembles, BayesNet) exhibit much
steeper MSE--Var fits ($b \gg 1$) and smaller increases in variance
($\Delta\text{Var}$), indicating sharper but less conservative uncertainty
even when they rank hard points reasonably well.

On the highly oscillatory sine shift, all methods degrade, but the
trade-offs differ. FDN preserves excellent ranking (Spearman $\rho$
near 1) and raises variance substantially OOD (large
$\Delta\mathrm{Var}$), yet its error grows faster than its variance (large $b$, large $\Delta\mathrm{MSE}$), yielding worse AURC. Deep ensembles show smaller
$\Delta\mathrm{MSE}$ and hence better AURC, but their ranking can be weaker. Overall, under matched capacity and update budgets, FDN’s main advantages are (i) calibrated scaling on smooth or piecewise-smooth shifts, where many baselines remain overconfident, and (ii) consistently high rank correlation across tasks, which makes FDN a strong triage signal even when absolute scale lags on rapidly oscillatory OOD. This highlights a clear avenue for improvement: stronger variance scaling on such shifts (e.g., temperature/flooring on $\sigma_\phi$, richer priors, or layer-wise $\beta$ schedules).

\noindent\textbf{Real regression benchmarks.}
On the real regression tasks (Airfoil, CCPP, Energy), the same
patterns largely persist (Figures~\ref{fig:real-msevar-grid} and
\ref{fig:bars-crps-vertical};
Tables~\ref{tab:airfoil-unified}--\ref{tab:energy-unified}).
FDN achieves reasonable in-distribution MSE and typically exhibits
large positive $\Delta\mathrm{Var}$, indicating that uncertainty
widens under feature-based shift. This comes with somewhat larger
$\Delta\mathrm{MSE}$ and $\Delta\mathrm{CRPS}$ than the sharpest
baselines on Airfoil, but more moderate values on CCPP and Energy, so
calibration degrades gradually rather than catastrophically. On
Airfoil, for example, FDN’s ID scatter lies close to the ideal
MSE$=$Var line and spreads out smoothly OOD, with strong Spearman
correlation; on CCPP and Energy the predictive variances remain
shift-aware and provide useful selective-risk behavior, even when the
absolute scale is not always better than the strongest baselines.
Taken together, the toy and UCI results suggest that input-conditioned
weight stochasticity is a viable and modular route to OOD-aware
regression: FDN behaves like a drop-in ``uncertainty layer’’ that can
be tuned via $\beta$ and hypernetwork capacity to trade off
in-distribution sharpness against OOD conservatism.

Additional qualitative diagnostics, including predictive means (Figure~\ref{fig:toy-mean-3x1}, \ref{fig:real-mean-3x1}), aggregated MSE--variance scatters (Figure~\ref{fig:toy-msevar-3x1}, \ref{fig:real-msevar-3x1}), and risk--coverage curves for all datasets (Figure~\ref{fig:toy-aurc-3x1}, \ref{fig:real-aurc-3x1}), are provided in (Appendix~\ref{app:baseline}).

\section{Limitations}
FDN’s input-conditioned \emph{weight} stochasticity, like other stochastic
layer designs, can overfit spurious cues if $\beta$ is too small or the
prior is too loose, making careful KL scheduling and prior choice
important. LP-FDN samples weights layer-by-layer, introducing additional
latency relative to a deterministic forward pass; test-time sampling
also incurs a compute–latency trade-off with the number of Monte Carlo
samples $K$, though we keep $K$ small in our experiments. Our study
focuses on low-dimensional regression with tabular datasets and modest
network backbones; extending FDN to high-dimensional inputs (e.g.,
images), deeper architectures, or structured outputs will likely require
additional engineering, such as low-rank or adapter-style hypernetworks
and tighter capacity control. On the calibration side, FDN can
under-scale variance in highly oscillatory OOD regimes (e.g., the sine
task), and the current design does not explicitly enforce
frequency-aware or spectral robustness.

\section{Conclusion}


We introduced Functional Distribution Networks (FDN), which amortize
input-conditioned distributions over weights to produce predictive
densities that remain sharp in-distribution and expand under shift.
Trained with a Monte Carlo objective and a $\beta$-weighted KL prior,
FDN achieves strong rank calibration across tasks and near-ideal scale
calibration on smooth and piecewise-smooth shifts, as measured by
Spearman correlation, MSE--Var fit, $\Delta\mathrm{Var}$,
$\Delta\mathrm{CRPS}$, and AURC. Under matched parameter, update, and
sampling budgets, FDN is competitive with Bayesian, ensemble, dropout,
and hypernetwork baselines on both 1D toy functions and UCI-style
regression benchmarks, while providing uncertainty suitable for
abstention and risk-aware inference.

Looking ahead, several directions are promising. On the modeling side,
stronger variance-scaling mechanisms (e.g., temperature or layer-wise
$\beta$ scheduling, structured priors) and frequency-aware conditioning
may reduce under-scaling on highly oscillatory OOD tasks. On the systems
side, adapter-style deployments that apply FDN to a small subset of
layers could inject uncertainty awareness into large-scale models with
modest overhead. Finally, extending FDN to classification, sequence
models, and structured prediction, or integrating it with other
uncertainty-aware modules (e.g., Neural Processes or diffusion-style
priors over weights), may enable a general toolkit for calibrated,
shift-aware deep learning.

\section*{Impact Statement}
This paper presents work whose goal is to advance the field of Machine Learning. There are many potential societal consequences of our work, none of which we feel must be specifically highlighted here. 

\bibliography{example_paper}
\bibliographystyle{icml2026}

\newpage
\appendix
\onecolumn


\section{Unified view via $q_\phi(\theta\mid x)$}
\label{app:q_phi}

All methods we consider can be written as
\[
p(y\mid x)\;=\;\int p(y\mid x,\theta)\;q_\phi(\theta\mid x)\,d\theta
\;\approx\;\frac{1}{K}\sum_{k=1}^K p\big(y\mid x,\theta^{(k)}\big),
\qquad 
\theta^{(k)}\sim q_\phi(\theta\mid x).
\]
In this paper, architectural layers are specified by the choice of $q_\phi(\theta\mid x)$. 
Framing models through $q_\phi(\theta\mid x)$ enables apples-to-apples comparisons: (i) how they set the spread of plausible weights, (ii) whether that spread adapts to the input, and (iii) how much compute they expend to form the predictive mixture. 
FDN’s module-level approach directly targets this knob: it provides local, input-aware uncertainty where it is inserted (e.g., the head or later blocks), broadens off-support as inputs drift from the training domain, and leaves the surrounding backbone and training loop unchanged.

\paragraph{FDN (IC/LP)}
\emph{FDN} makes $q_\phi$ \textbf{input-conditional and stochastic}.
A common choice is diagonal-Gaussian, factorized by layer:
\[
q_\phi(\theta\mid x)=\prod_{\ell}\mathcal{N}\!\big(\mu_\ell(x),\operatorname{diag}\sigma_\ell^2(x)\big).
\]
This \textbf{input-conditional} variant is \emph{IC-FDN}. For \textbf{layer-propagated} conditioning (\emph{LP-FDN}), the $\ell$-th layer’s weight distribution depends \emph{only} on the previous activation:
\[
q_{\phi}(\theta_\ell\mid x)
=\mathcal{N}\!\big(\mu_\ell(a_{\ell-1}),\operatorname{diag}\sigma_\ell^2(a_{\ell-1})\big),
\qquad a_0:=x,
\]
and sampling proceeds sequentially across layers along the same Monte Carlo sample path. 
This induces a first-order Markov structure in depth, allowing uncertainty to expand as signals propagate—later layers can broaden even when early layers remain sharp. 
We regularize with a per-layer KL:
\[
\beta\sum_{\ell=1}^{L}\displaystyle KL ( q_\phi(\theta_{\ell}\mid x) \Vert p_0(\theta_{\ell}) ).
\]
More generally, one could condition longer histories $a_{0:\ell-1}$; in this paper, we restrict to first-order (one-step) conditioning. Note, in the limit $\sigma_\ell\to 0$ for all $\ell$, the model collapses to a deterministic layer-conditioned Hypernetwork.

\paragraph{Deterministic Hypernetwork.}
A deterministic Hypernetwork $G_\phi$ maps the input to weights, yielding a \textbf{degenerate} $q$:
\[
q_\phi(\theta\mid x)=\delta\!\big(\theta - G_\phi(x)\big),
\qquad 
p(y\mid x)=p\!\big(y\mid x,G_\phi(x)\big).
\]
Training typically uses NLL or MSE; weight decay on $\phi$ can be interpreted as a MAP prior on the Hypernetwork parameters. 
Because $q_\phi$ is a Dirac-Delta, there is no weight-space uncertainty: any predictive uncertainty must come from the observation model (e.g., a heteroscedastic head) or post-hoc calibration. 
Compared to stochastic variants, this adds no KL term and no MC averaging, but can increase per-example compute due to generating weights via $G_\phi$.

\paragraph{Gaussian HyperNetwork}
A \emph{Stochastic} Hypernetwork outputs a \textbf{global} posterior (or context-only):
\[
q_\phi(\theta\mid x)\equiv q_\phi(\theta\mid h)=\prod_{\ell}\mathcal{N}(\mu_{\ell}(h),\operatorname{diag}\sigma_{\ell}^{2}(h)),
\]
i.e., independent of the query $x$ (but dependent on a learnable latent task vector $h$). This is variational BNN with parameters produced by a Hypernetwork.

\paragraph{Bayesian Neural Network (Bayes-by-Backprop).}
A standard variational BNN uses an \textbf{$x$-independent} approximate posterior:
\[
q_\phi(\theta\mid x)\equiv q_\phi(\theta)
=\prod_{\ell}\mathcal{N}\!\big(\mu_\ell,\operatorname{diag}\sigma_\ell^2\big),
\]
and the same $\beta$-ELBO objective with closed-form diagonal-Gaussian KL. Because $q_\phi(\theta | x)$ is global, predictive uncertainty does not adapt to $x$ except via the likelihood term, which can under-react off-support compared to input-conditional alternatives. On the other hand, the objective is simple and sampling cost is amortized across inputs, though matching ensemble-like diversity typically requires larger posterior variances or multiple posterior samples at test time.

\paragraph{MLP with Dropout.}
MLP with dropout induces a distribution over effective weights via random masks $m$:
\[
q_\phi(\theta\mid x)\equiv q_\phi(\theta)\quad\text{(implicit via dropout masks, independent of $x$)},
\]
and inference averages predictions over sampled masks \citep{gal2016dropout}. 

\paragraph{Deep Ensembles.}
An $M$-member ensemble of MLPs corresponds to a \textbf{finite mixture of deltas}:
\[
q(\theta\mid x)\;\equiv\;\frac{1}{M}\sum_{m=1}^{M}\delta\!\big(\theta-\theta_m\big),
\qquad
p(y\mid x)\;=\;\frac{1}{M}\sum_{m=1}^{M} p\big(y\mid x,\theta_m\big),
\]
where each $\theta_m$ is trained independently from a different initialization (and typically a different data order/augmentation). 
There is no explicit KL regularizer; diversity arises implicitly from independent training trajectories. 
Inference cost scales linearly with $M$ (one forward pass per member), and for fair comparisons we match total update or compute budgets by reducing epochs.


\section{Training objective}
\label{app:training}
FDN is amortized variational inference with the latent weights $\theta$ and input-conditional posterior $q_\phi(\theta\mid x)$ realized by small hypernetworks via the reparameterization
$\theta^{(k)} = g_\phi(x,\varepsilon^{(k)})$ with $\varepsilon^{(k)}\!\sim\!\mathcal N(0,I)$.
For a single $(x,y)$ the ELBO is
\begin{equation}
\log p(y\mid x)\ \ge\ 
\underbrace{\mathbb{E}_{q_\phi(\theta\mid x)}\big[\log p(y\mid x,\theta)\big]}_{\text{data term}}
\ -\
\underbrace{KL\!\big(q_\phi(\theta\mid x)\,\|\,p_0(\theta)\big)}_{\text{regularizer}},
\end{equation}
with a simple prior $p_0(\theta)=\prod_\ell \mathcal N(0,\sigma_0^2 I)$.

\paragraph{(A) $\beta$–ELBO (mean of logs).}
We minimize the negative $\beta$–ELBO with $K$ Monte Carlo draws:
\begin{equation}
\mathcal L_{\beta\text{-ELBO}}
= -\frac{1}{K}\sum_{k=1}^{K}\log p\!\big(y\mid x,\theta^{(k)}\big)
\;+\; \beta\, KL\!\big(q_\phi(\theta\mid x)\,\|\,p_0(\theta)\big),
\qquad \theta^{(k)}\!\sim q_\phi(\theta\mid x).
\end{equation}
Here $\beta{=}1$ recovers standard VI; $\beta\!\neq\!1$ implements capacity control / tempered VI \citep{higgins2017beta,alemi2017deep,dziugaite2017computing}. We use simple warm-ups for $\beta$ early in training.

\paragraph{(B) IWAE variant (log of means; tighter bound).}
As a reference, the importance-weighted bound is
\begin{equation}
\mathcal L_{\text{IWAE}}
= -\log\!\left(\frac{1}{K}\sum_{k=1}^{K}
\frac{p_0(\theta^{(k)})\,p(y\mid x,\theta^{(k)})}{q_\phi(\theta^{(k)}\mid x)}\right),
\qquad \theta^{(k)}\!\sim q_\phi(\theta\mid x),
\end{equation}
which implicitly accounts for the KL via the weights and typically needs no extra $\beta$ \citep{burda2016importance}. We report main results with (A) for simplicity and stability.

\paragraph{KL decomposition (IC vs.\ LP).}
For \emph{IC-FDN}, layer posteriors condition directly on $x$, so the KL sums over layers and averages over the minibatch.
For \emph{LP-FDN}, layer $\ell$ conditions on a sampled hidden state $a_{\ell-1}^{(k)}(x)$; the KL is therefore averaged over this upstream randomness:
\[
KL\big(q_{\phi,\ell}(\theta_\ell\!\mid a_{\ell-1}^{(k)}(x))\ \|\ p_0\big)\quad\text{with}\quad
\mathbb E_k[\cdot]\ \text{across samples }k.
\]
With diagonal Gaussians, each layer’s closed-form term is
\[
KL\!\left(\mathcal N(\mu,\mathrm{diag}\,\sigma^2)\,\|\,\mathcal N(0,\sigma_0^2 I)\right)
= \tfrac{1}{2}\sum_j\!\left(\frac{\sigma_j^2+\mu_j^2}{\sigma_0^2}-1-\log\frac{\sigma_j^2}{\sigma_0^2}\right),
\]
and we implement the variance floor via $\sigma=\varepsilon+\mathrm{softplus}(\rho)$ (no hard clamp). 

\textit{Remark.} Future work should investigate \emph{layer-specific} $\beta$ schedules to control where uncertainty is expressed across depth (e.g., larger $\beta$ in early layers for stability, smaller $\beta$ near the output to permit output-scale variance), with the aim of tightening scale calibration ($b\!\to\!1$, $a\!\to\!0$) and improving AURC/CRPS under oscillatory OOD.

\section{Notation and symbols}
\label{app:notation}

\FloatBarrier

\begin{table}[H]
  \centering
  \caption{Key symbols, parameters, and dimensions used throughout the paper.}
  \label{tab:notation-core}
  \begingroup
  \small
  \setlength{\tabcolsep}{6pt}
  \begin{tabular}{lll}
    \toprule
    \textbf{Symbol} & \textbf{Description} & \textbf{Dimension / Type} \\
    \midrule
    $x$                  & Input / covariate                         & $\mathbb{R}^{d_x}$ \\
    $y$                  & Target / response                         & $\mathbb{R}^{d_y}$ \\
    $\mathcal{D}$        & Training dataset                          & $\{(x_i,y_i)\}_{i=1}^N$ \\
    $N$                  & Number of training examples               & $\mathbb{N}$ \\
    $T$                  & Number of test examples                   & $\mathbb{N}$ \\
    $d_x$                & Input dimension                           & scalar \\
    $d_y$                & Output dimension                          & scalar \\
    $D_y$                & Gaussian head output dim.                 & $\frac{d_y(d_y+3)}{2}$ \\
    \midrule
    $f_\theta$           & Base network (predictive head)            & $f_\theta:\mathbb{R}^{d_x}\!\to\!\mathbb{R}^{D_y}$ \\
    $\theta$             & All base-network weights (all layers)     & $\mathbb{R}^{P}$ \\
    $\phi$               & FDN / variational / hypernetwork params   & parameter vector \\
    $p(y\mid x,\theta)$  & Likelihood (Gaussian head)                & $\mathcal{N}(y;\mu_\theta(x),\Sigma_\theta(x))$ \\
    $p_0(\theta)$        & Weight prior                              & $\mathcal{N}(0,\sigma_0^2 I)$ \\
    $q_\phi(\theta\mid x)$ & Input-conditioned weight posterior      & $\mathcal{N}(\mu_\phi(x),\operatorname{diag}\sigma_\phi^2(x))$ \\
    $\beta$              & KL weight in $\beta$--ELBO                & scalar $\ge 0$ \\
    $\mathcal{L}_{\beta\text{-ELBO}}$ & Training loss (per-example)  & scalar \\
    \midrule
    $L$                  & Number of layers in base net              & scalar \\
    $d_\ell$             & Width of layer $\ell$                     & scalar \\
    $s_\ell$             & Conditioning signal for layer $\ell$      & $\mathbb{R}^{d_{s,\ell}}$ \\
    $d_{s,\ell}$         & Dimension of $s_\ell$                     & $d_x$ (IC) or $d_{\ell-1}$ (LP) \\
    $A_\ell$             & Per-layer hypernetwork                    & $A_\ell:\mathbb{R}^{d_{s,\ell}}\!\to\!\mathbb{R}^{P_\ell}$ \\
    $P_\ell$             & \# Gaussian params for layer $\ell$       & $2(d_{\ell-1}d_\ell + d_\ell)$ \\
    $P$                  & Total \# trainable parameters             & $\sum_{\ell} P_\ell$ (plus head) \\
    $h_{\mathrm{hyp}}$   & Hypernetwork hidden width                 & scalar \\
    \midrule
    $K$                  & MC samples per input (train/test)         & $\mathbb{N}$ \\
    $M$                  & Ensemble size (DeepEnsembleNet)           & $\mathbb{N}$ \\
    $\sigma_0$           & Prior std.\ for weights                   & scalar $>0$ \\
    $\sigma^2$           & Observation noise variance (homoscedastic)& scalar $>0$ \\
    $\varepsilon$        & Variance floor in $\sigma=\varepsilon+\mathrm{softplus}(\rho)$ & scalar $>0$ \\
    $\widehat{\mu}(x)$   & Predictive mean                           & $\mathbb{R}^{d_y}$ \\
    $\widehat{\mathrm{Var}}[Y\mid x]$ & Predictive variance estimator & scalar (for $d_y{=}1$) \\
    \midrule
    $\mathrm{MSE}$       & Mean squared error                        & scalar \\
    $\mathrm{CRPS}$      & Continuous ranked prob.\ score            & scalar \\
    $\mathrm{NLL}$       & Neg.\ log predictive density              & scalar \\
    $\rho$               & Spearman rank correlation                 & scalar $\in[-1,1]$ \\
    $\mathrm{AURC}$      & Area under risk--coverage curve           & scalar \\
    $\Delta(\cdot)$      & ID$\!\to$OOD metric delta                 & $\mathbb{E}_{\text{OOD}}[\cdot] - \mathbb{E}_{\text{ID}}[\cdot]$ \\
    \bottomrule
  \end{tabular}
  \endgroup
\end{table}

\section{Heteroscedastic likelihood and variance decomposition}
\label{app:hetero}

We briefly collect the forms of the Gaussian $\beta$--ELBO used in this work
and the associated decomposition of predictive variance into epistemic and
aleatoric components.

\paragraph{General Gaussian head.}
For a Gaussian predictive head with possibly heteroscedastic, full-covariance
noise, the per-example $\beta$--ELBO for datum $(x_i,y_i)$ is
\begin{align}
\mathcal{L}_{\text{Gauss}}^{(i)}
&=
\frac{1}{2K}\sum_{k=1}^{K}
\Big[
\big\|y_i-\mu_{\theta_k}(x_i)\big\|^{2}_{\left(\Sigma_{\theta_k}(x_i)\right)^{-1}}
\;+\; \log\det\!\big(2\pi\,\Sigma_{\theta_k}(x_i)\big)
\Big]
\;+\;
\beta\,KL\!\big( q_\phi(\theta\mid x_i) \,\Vert\, p_0(\theta) \big),
\label{eq:gauss-elbo-appendix}
\end{align}

where $\|v\|^{2}_{A}:=v^\top A v$.
In the \emph{homoscedastic} case the covariance is constant across
inputs, so a full $\Sigma$ encodes a single, global correlation structure
among output dimensions; if $\Sigma$ is diagonal, the data term reduces
to a (constant-)weighted least-squares plus a constant log-determinant.
In contrast, \emph{heteroscedastic} models use an input-dependent
$\Sigma_\theta(x)$, so the weights (and, for full $\Sigma_\theta(x)$,
correlations) vary with $x$ and with the sampled weights~$\theta$.

For $d_y=1$ the covariance reduces to a scalar $\sigma_{\theta}^{2}(x)$.
We say the noise is \emph{homoscedastic in $x$} if
$\sigma_{\theta}^{2}(x)\equiv\sigma_{\theta}^{2}$ (constant across
inputs for a fixed $\theta$), and \emph{heteroscedastic in $x$} if
$\sigma_{\theta}^{2}(x)$ varies with $x$.
Orthogonally, because we draw stochastic weights $\theta^{(k)}$, one
can distinguish dependence on the sampled weights:
\emph{homoscedastic in $\theta$} means $\sigma_{\theta}^{2}(x)$ is
effectively deterministic (identical across $\theta^{(k)}$ for a given
$x$), while \emph{heteroscedastic in $\theta$} means
$\sigma_{\theta^{(k)}}^{2}(x)$ changes with the sampled weights (as in
FDN/BNN where the variance head depends on $\theta^{(k)}$).

In the \textbf{isotropic heteroscedastic} case,
$\Sigma_{\theta^{(k)}}(x)=\sigma^{2}_{\theta^{(k)}}(x) I$, the Gaussian
negative log-likelihood (NLL) contribution is
\begin{align}
\frac{1}{2K}\sum_{k=1}^{K}\!\Bigg[
\left(\frac{y_i - \mu_{\theta^{(k)}}(x_i)}{\sigma_{\theta^{(k)}}(x_i)}\right)^{2}
\;+\; \log\!\big(2\pi\,\sigma^{2}_{\theta^{(k)}}(x_i)\big)
\Bigg],
\end{align}
i.e., a \emph{per-input, per-sample weighted} MSE plus a variance penalty.
If one instead assumes \textbf{isotropic homoscedastic} noise
($\sigma^2$ constant), the data term is proportional to the MSE up to an
additive constant; in practice the fit--regularization trade-off can be
tuned either by setting $\sigma^2$ or, equivalently, by adjusting
$\beta$ to re-balance the data term against the KL.

Since our main focus is on uncertainty-aware metrics rather than
cross-output correlations, in the experiments we restrict attention to
$d_y=1$ and use the isotropic homoscedastic case, with a single constant
variance $\sigma^2$ that we absorb into $\beta$.

\paragraph{Scalar homoscedastic $\beta$--ELBO.}
Consider the latent-weight model
\[
\theta \sim p_0(\theta),\qquad
y \mid x,\theta \sim \mathcal N\!\big(f_\theta(x),\,\sigma^2\big),
\]
with a variational family $q_\phi(\theta\mid x)$ (IC-/LP-FDN).
For one datum $(x_i,y_i)$ the standard ELBO is
\[
\log p(y_i\mid x_i)\ \ge\
\underbrace{\mathbb{E}_{q_\phi}\!\big[\log p(y_i\mid x_i,\theta)\big]}_{\text{data term}}
\;-\;
\underbrace{KL\!\big(q_\phi(\theta\mid x_i)\,\|\,p_0(\theta)\big)}_{\text{regularizer}}.
\]
Using the Gaussian likelihood,
\[
\log p(y_i\mid x_i,\theta)
= -\frac{1}{2\sigma^2}\big(y_i-f_\theta(x_i)\big)^2\;-\;\tfrac{1}{2}\log(2\pi\sigma^2).
\]
Plugging into the bound and negating yields the per-example loss
\[
\mathcal L_{\text{ELBO}}^{(i)}
= \frac{1}{2\sigma^2}\ \mathbb{E}_{q_\phi(\theta\mid x_i)}\!\big[(y_i-f_\theta(x_i))^2\big]
\;+\;KL\!\big(q_\phi(\theta\mid x_i)\,\|\,p_0(\theta)\big)
\;+\;\tfrac{1}{2}\log(2\pi\sigma^2).
\]
Using $K$ reparameterized samples $\theta^{(k)}\!\sim q_\phi(\theta\mid x_i)$
gives the unbiased Monte Carlo estimator
\[
\boxed{\;
\mathcal L_{\text{ELBO}}^{(i)}
\approx
\frac{1}{2K\sigma^2}\sum_{k=1}^K \big(y_i - f_{\theta^{(k)}}(x_i)\big)^2
\;+\;KL\!\big(q_\phi(\theta\mid x_i)\,\|\,p_0(\theta)\big)
\;+\;\tfrac{1}{2}\log(2\pi\sigma^2)\; }.
\]
Since $\tfrac{1}{2}\log(2\pi\sigma^2)$ does not depend on $\phi$ or
$\theta$, it can be dropped during optimization.
If $\sigma^2$ is fixed, the data term is just a rescaled MSE, so
\[
(2\sigma^2)\,\mathcal L_{\text{ELBO}}^{(i)} \doteq
\frac{1}{K}\sum_{k=1}^K \big(y_i - f_{\theta^{(k)}}(x_i)\big)^2
\;+\;\underbrace{(2\sigma^2)}_{\beta}\,KL\!\big(q_\phi(\theta\mid x_i)\,\|\,p_0(\theta)\big),
\]
showing that choosing constant $\sigma^2$ is equivalent to training with
a $\beta$--ELBO, with $\beta$ simply rescaling the effective KL weight
(capacity control). In this paper we directly use a $\beta$--ELBO for
training.

\paragraph{Heteroscedastic observation model (scalar case).}
If we allow the observation variance to depend on $x$ and the sampled
weights $\theta$,
\[
Y \mid x,\theta \sim \mathcal N\!\big(f_\theta(x),\,\sigma_{\theta}^{2}(x)\big)
\qquad (d_y=1),
\]
the per-example $\beta$--ELBO becomes
\[
\mathcal L_{\text{het}}^{(i)}
=\frac{1}{2K}\sum_{k=1}^{K}
\Bigg[
\frac{\big(y_i - f_{\theta^{(k)}}(x_i)\big)^2}{\sigma_{\theta^{(k)}}^{2}(x_i)}
\;+\; \log\!\big(2\pi\,\sigma_{\theta^{(k)}}^{2}(x_i)\big)
\Bigg]
\;+\;
\beta\,KL\!\big(q_\phi(\theta\mid x_i)\,\|\,p_0(\theta)\big),
\]
i.e., a weighted least-squares (WLS) term plus a variance penalty, with
weights $w^{(k)}(x_i)=1/\sigma_{\theta^{(k)}}^{2}(x_i)$ learned jointly
with the mean.

\textit{Parameterization and stability.}
We parameterize
\[
\sigma_{\theta}(x) \;=\; \varepsilon + \operatorname{softplus}\big(\rho_{\theta}(x)\big),\qquad \varepsilon=10^{-3},
\]
which guarantees positivity and avoids numerical collapse.
To mitigate variance blow-up in early training, one can
(i) apply gentle weight decay on $\rho_{\theta}$,
(ii) clip $s_{\theta}(x)=\log\sigma_{\theta}^{2}(x)$ to a reasonable
range, or (iii) use a short $\beta$ warm-up so the likelihood term
dominates initially.

\paragraph{Predictive variance decomposition.}
Let $\theta\!\sim\!q_\phi(\theta\mid x)$ and, given $(x,\theta)$,
\[
Y \mid x,\theta \;\sim\; \mathcal N\!\big(\mu_\theta(x),\,\sigma^2_\theta(x)\big).
\]

\emph{Scalar case.}
The predictive (marginal) variance decomposes as
\begin{equation}
\mathrm{Var}[Y\mid x]
\;=\;
\underbrace{\mathbb{E}_{\theta\sim q_\phi}\!\big[\sigma^2_\theta(x)\big]}_{\text{aleatoric}}
\;+\;
\underbrace{\mathrm{Var}_{\theta\sim q_\phi}\!\big[\mu_\theta(x)\big]}_{\text{epistemic}}.
\label{eq:var-decomp-scalar}
\end{equation}

\emph{Proof.}
By the law of total expectation,
$\mathbb{E}[Y\mid x]=\mathbb{E}_\theta[\mathbb{E}[Y\mid x,\theta]]
=\mathbb{E}_\theta[\mu_\theta(x)]$.
By the law of total variance,
\[
\mathrm{Var}[Y\mid x]
=\mathbb{E}_\theta\!\big[\mathrm{Var}(Y\mid x,\theta)\big]
+\mathrm{Var}_\theta\!\big(\mathbb{E}[Y\mid x,\theta]\big)
=\mathbb{E}_\theta\!\big[\sigma^2_\theta(x)\big]
+\mathrm{Var}_\theta\!\big[\mu_\theta(x)\big].\quad\square
\]

\emph{Vector-output version.}
For $Y\in\mathbb{R}^{d_y}$ with
$Y\mid x,\theta\sim \mathcal N(\mu_\theta(x),\Sigma_\theta(x))$ the
predictive covariance is
\begin{equation}
\operatorname{Cov}[Y\mid x]
\;=\;
\underbrace{\mathbb{E}_{\theta}\!\big[\Sigma_\theta(x)\big]}_{\text{aleatoric}}
\;+\;
\underbrace{\operatorname{Cov}_{\theta}\!\big[\mu_\theta(x)\big]}_{\text{epistemic}},
\label{eq:var-decomp-vector}
\end{equation}
obtained by the matrix form of the law of total variance.

\paragraph{Monte Carlo estimators.}
With samples $\theta^{(k)}\!\sim q_\phi(\theta\mid x)$ we estimate the
predictive mean and epistemic variance as
\[
\hat\mu(x)=\frac{1}{K}\sum_{k=1}^K \mu_{\theta^{(k)}}(x),\qquad
\widehat{\mathrm{Var}}_{\text{epi}}(x)=\frac{1}{K}\sum_{k=1}^K\big(\mu_{\theta^{(k)}}(x)-\hat\mu(x)\big)^2,
\]
and obtain the total predictive variance via the decomposition
\eqref{eq:var-decomp-scalar},
\[
\widehat{\mathrm{Var}}[Y\mid x]
=\frac{1}{K}\sum_{k=1}^K \sigma^2_{\theta^{(k)}}(x)
\;+\; \widehat{\mathrm{Var}}_{\text{epi}}(x).
\]
For $d_y>1$, replace squared deviations by outer products to estimate
covariances, in accordance with \eqref{eq:var-decomp-vector}.

In our experiments we use the homoscedastic scalar case
($\sigma_\theta^2(x)\equiv\sigma^2$), so the aleatoric variance reduces
to a constant and all input-dependent variability in
$\mathrm{Var}[Y\mid x]$ comes from the epistemic component
$\mathrm{Var}_\theta[\mu_\theta(x)]$; the heteroscedastic extension
above is included for completeness.

\section{FDN Training and Prediction Algorithm}
\label{app:alg}

\FloatBarrier

\begin{algorithm}[H]
\caption{FDN (Unified for IC-/LP-FDN): Training and Prediction}
\label{alg:fdn-unified}
\begin{algorithmic}[1]
\STATE \textbf{Inputs:} dataset $\mathcal D=\{(x_i,y_i)\}_{i=1}^N$; base net $f_\theta$ with layers $1{:}L$; per-layer samplers $q_\phi^l(\theta_l\!\mid\!c_l)$ (diag. Gaussians); prior $p_0(\theta)=\prod_l p_0^l(\theta_l)$; MC $K$; KL schedule $\{\beta_t\}$; variant $v\in\{\text{IC},\text{LP}\}$.
\FOR{step $t=1,2,\dots$}
  \STATE Sample minibatch $\mathcal B$; set $\Sigma_{\mathrm{NLL}}\!\gets 0$, $\Sigma_{\mathrm{KL}}\!\gets 0$
  \FOR{each $(x,y)\in\mathcal B$}
    \FOR{$k=1,\dots,K$}
      \STATE $h_0^{(k)}\!\gets x$
      \FOR{$l=1,\dots,L$}
        \STATE $c_l \gets \begin{cases}
          x & \text{if } v=\text{IC}\\
          h_{l-1}^{(k)} & \text{if } v=\text{LP}
        \end{cases}$
        \STATE $\varepsilon_l^{(k)}\!\sim\!\mathcal N(0,I)$, \;\; $\theta_l^{(k)}\!\leftarrow\!\mu_\phi^l(c_l)+\sigma_\phi^l(c_l)\odot \varepsilon_l^{(k)}$
        \STATE $h_l^{(k)}\!\gets \text{layer}_l\!\big(h_{l-1}^{(k)};\theta_l^{(k)}\big)$
        \STATE $\Sigma_{\mathrm{KL}} \mathrel{+}= \displaystyle KL \big(q_\phi^l(\theta_l\!\mid\!c_l)\,\|\,p_0^l(\theta_l)\big)$
      \ENDFOR
      \STATE $(\mu^{(k)},\Sigma^{(k)}) \gets \text{head}\!\big(h_L^{(k)}\big)$ \COMMENT{$\Sigma^{(k)}$ may be fixed (homoscedastic)}
      \STATE $\ell_k \gets \log \mathcal N\!\big(y;\mu^{(k)},\Sigma^{(k)}\big)$
    \ENDFOR
    \STATE $\Sigma_{\mathrm{NLL}} \mathrel{+}= -\tfrac{1}{K}\sum_{k=1}^K \ell_k$ \COMMENT{ELBO (mean-of-logs)}
  \ENDFOR
  \STATE $\mathcal L \gets \tfrac{1}{N}\Big(\Sigma_{\mathrm{NLL}} + \beta_t\,\Sigma_{\mathrm{KL}}\Big)$; \quad update $\phi$ by backprop on $\mathcal L$
\ENDFOR
\STATE \textbf{Predict at $x_\star$:} repeat the per-layer sampling with $c_l{=}\{x_\star \text{ or } h_{l-1}^{(k)}\}$ per $v$ to obtain $(\mu_\star^{(k)},\Sigma_\star^{(k)})$, and return $\hat p(y\!\mid\!x_\star)\!\approx\!\tfrac{1}{K}\!\sum_k \mathcal N\!\big(y;\mu_\star^{(k)},\Sigma_\star^{(k)}\big)$.
\end{algorithmic}
\end{algorithm}

\clearpage 

\section{Supplementary Figures and Tables}
\label{app:baseline}
\begin{figure*}
  \centering
  \captionsetup[subfigure]{justification=centering,singlelinecheck=false,skip=2pt}

  \begin{subfigure}[t]{0.96\textwidth}
    \centering
    \includegraphics[width=0.95\linewidth]{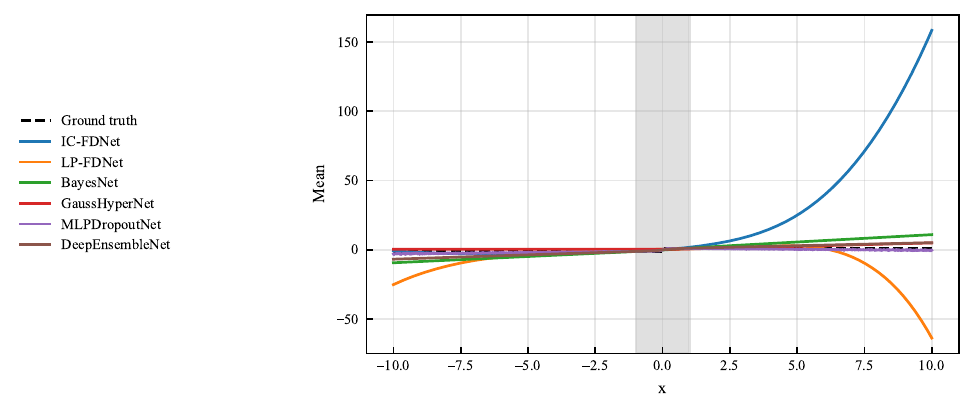}
    \subcaption{Step: $H(x)$}
    \label{fig:toy-mean-step}
  \end{subfigure}

  \vspace{0.4em}

  \begin{subfigure}[t]{0.96\textwidth}
    \centering
    \includegraphics[width=0.95\linewidth]{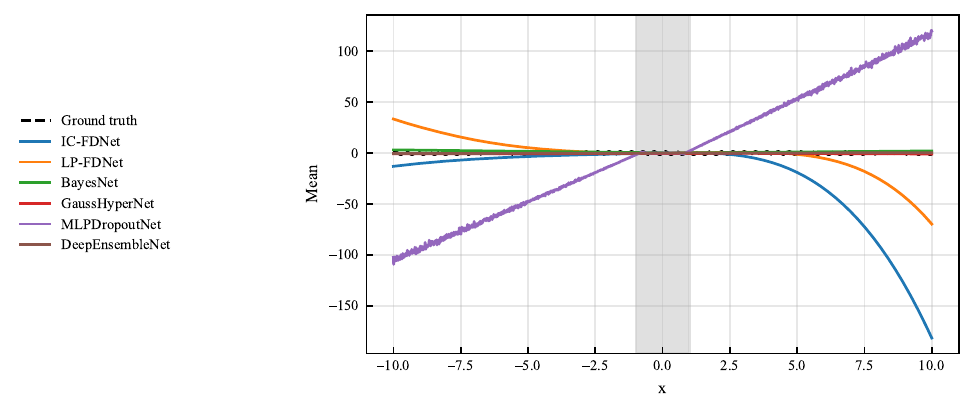}
    \subcaption{Sine: $1.54\,\sin(2.39\,x)$}
    \label{fig:toy-mean-sine}
  \end{subfigure}

  \vspace{0.4em}

  \begin{subfigure}[t]{0.96\textwidth}
    \centering
    \includegraphics[width=0.95\linewidth]{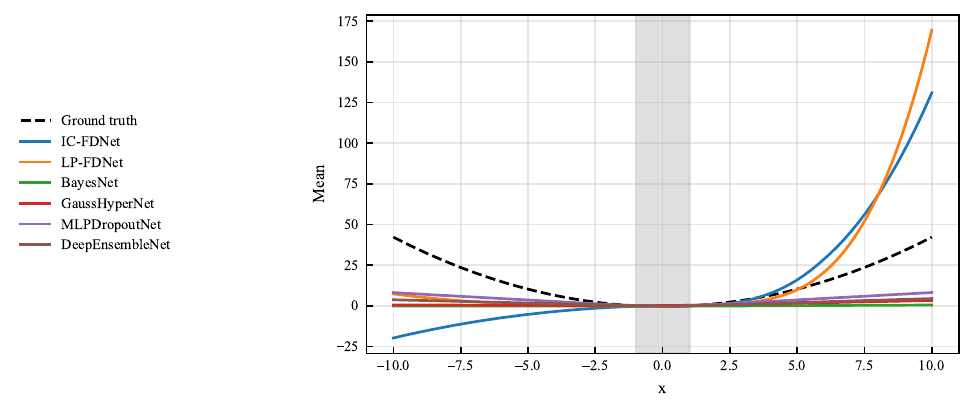}
    \subcaption{Quadratic: $0.43x^{2}-0.41$}
    \label{fig:toy-mean-quad}
  \end{subfigure}

  \caption{Predictive mean vs.\ input $x$ for the three synthetic 1D toy tasks
  (step, sine, quadratic), with the ground-truth function overlaid. Shaded region corresponds to the interpolation/ ID points.}
  \label{fig:toy-mean-3x1}
\end{figure*}

\begin{figure*}
  \centering
  \captionsetup[subfigure]{justification=centering,singlelinecheck=false,skip=2pt}

  \begin{subfigure}[t]{0.96\textwidth}
    \centering
    \includegraphics[width=0.95\linewidth]{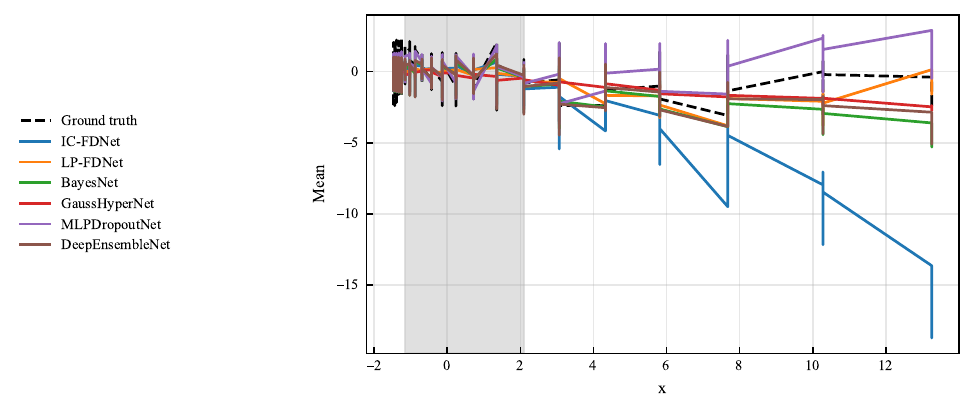}
    \subcaption{Airfoil Self-Noise}
    \label{fig:real-mean-airfoil}
  \end{subfigure}

  \vspace{0.4em}

  \begin{subfigure}[t]{0.96\textwidth}
    \centering
    \includegraphics[width=0.95\linewidth]{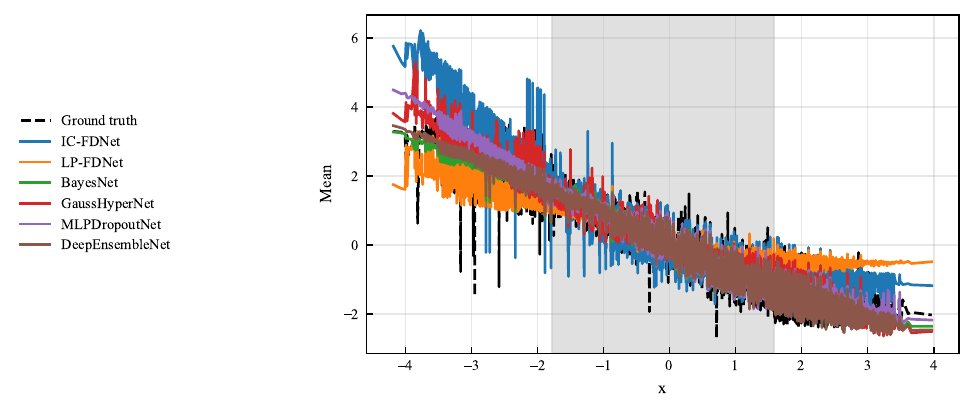}
    \subcaption{CCPP Power Plant}
    \label{fig:real-mean-ccpp}
  \end{subfigure}

  \vspace{0.4em}

  \begin{subfigure}[t]{0.96\textwidth}
    \centering
    \includegraphics[width=0.95\linewidth]{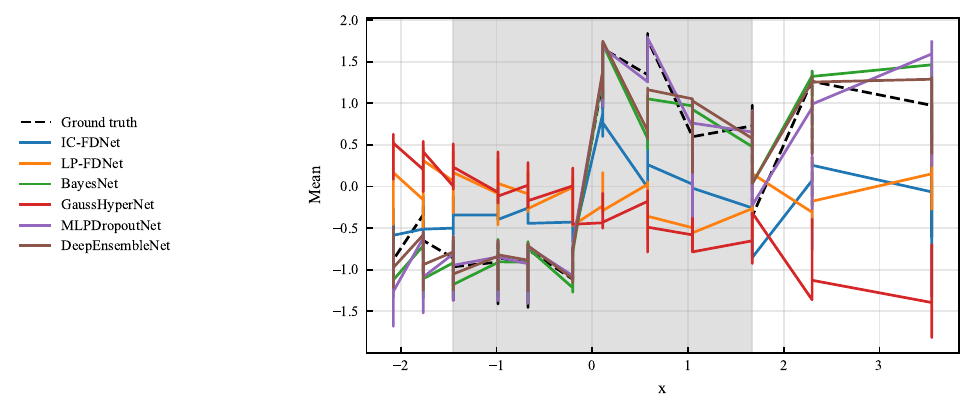}
    \subcaption{Energy Efficiency}
    \label{fig:real-mean-energy}
  \end{subfigure}

  \caption{Predictive mean vs.\ standardized split feature $x$ for the three real
  regression datasets (Airfoil Self-Noise, CCPP Power Plant, Energy Efficiency). Shaded region corresponds to the interpolation/ ID points.}
  \label{fig:real-mean-3x1}
\end{figure*}

\begin{figure*}
  \centering
  \captionsetup[subfigure]{justification=centering,singlelinecheck=false,skip=2pt}

  \begin{subfigure}[t]{0.96\textwidth}
    \centering
    \includegraphics[width=0.95\linewidth]{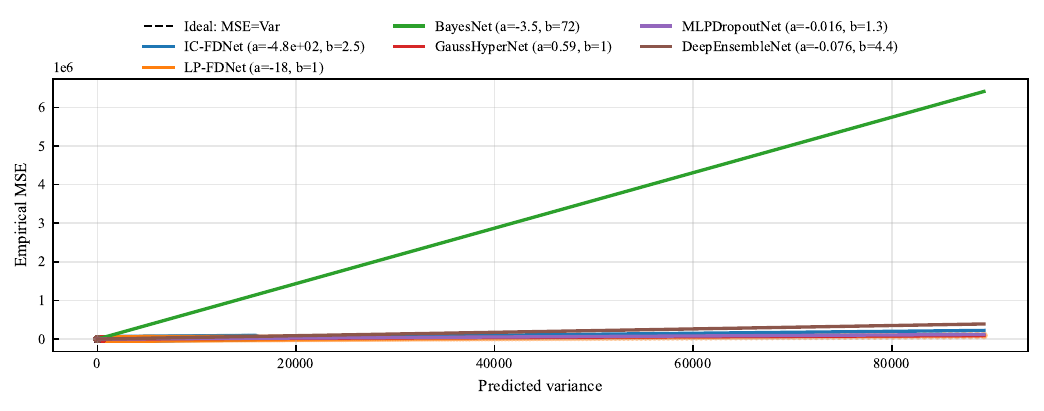}
    \subcaption{Step: $H(x)$}
    \label{fig:toy-msevar-step}
  \end{subfigure}

  \vspace{0.4em}

  \begin{subfigure}[t]{0.96\textwidth}
    \centering
    \includegraphics[width=0.95\linewidth]{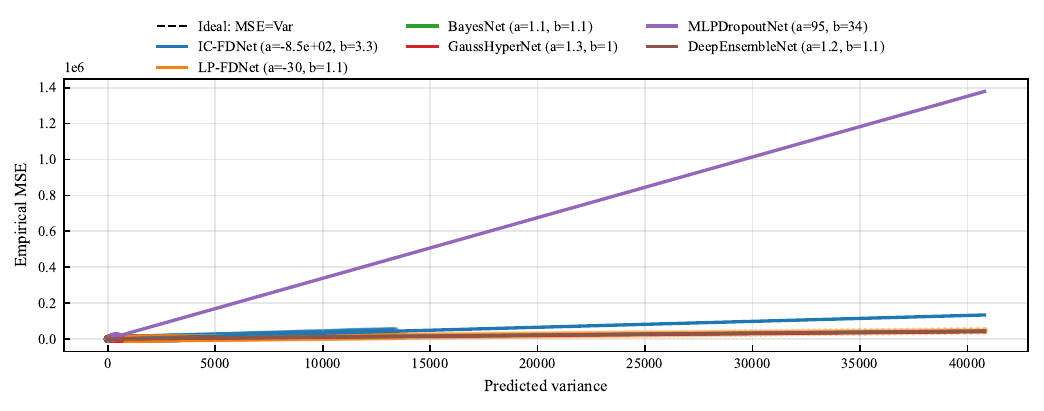}
    \subcaption{Sine: $1.54\,\sin(2.39\,x)$}
    \label{fig:toy-msevar-sine}
  \end{subfigure}

  \vspace{0.4em}

  \begin{subfigure}[t]{0.96\textwidth}
    \centering
    \includegraphics[width=0.95\linewidth]{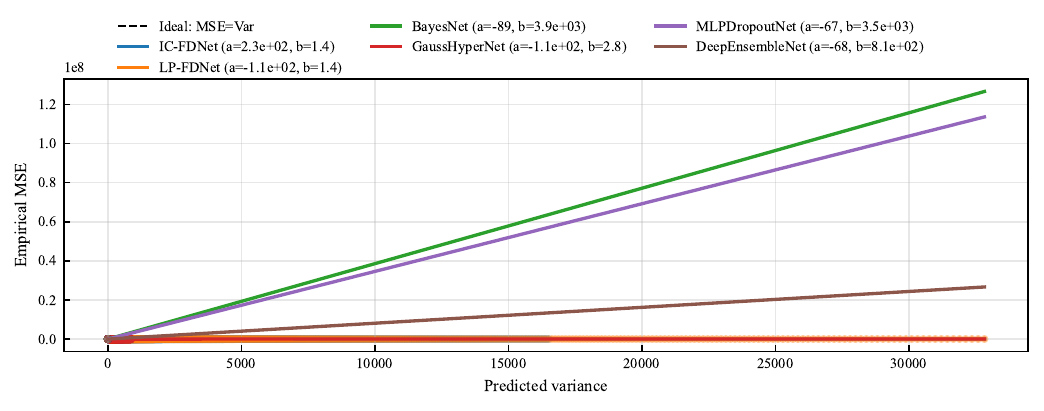}
    \subcaption{Quadratic: $0.43x^{2}-0.41$}
    \label{fig:toy-msevar-quad}
  \end{subfigure}

  \caption{MSE vs.\ predicted variance scatter plots for the three toy tasks.
  Each panel aggregates ID and OOD test points; the dashed line shows the ideal
  calibration $\mathrm{MSE}=\mathrm{Var}$. Legends in the PDFs report Spearman’s
  $\rho$ and linear-fit slope/intercept.}
  \label{fig:toy-msevar-3x1}
\end{figure*}

\begin{figure*}
  \centering
  \captionsetup[subfigure]{justification=centering,singlelinecheck=false,skip=2pt}

  \begin{subfigure}[t]{0.96\textwidth}
    \centering
    \includegraphics[width=0.95\linewidth]{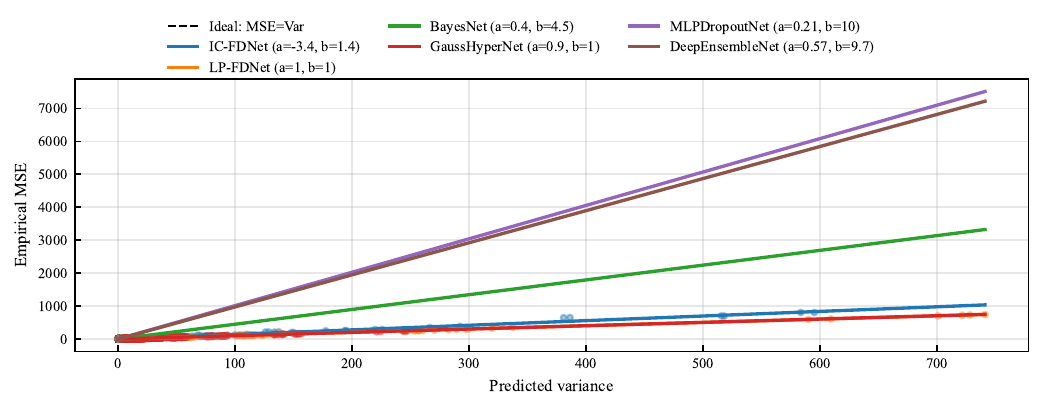}
    \subcaption{Airfoil Self-Noise}
    \label{fig:real-msevar-airfoil}
  \end{subfigure}

  \vspace{0.4em}

  \begin{subfigure}[t]{0.96\textwidth}
    \centering
    \includegraphics[width=0.95\linewidth]{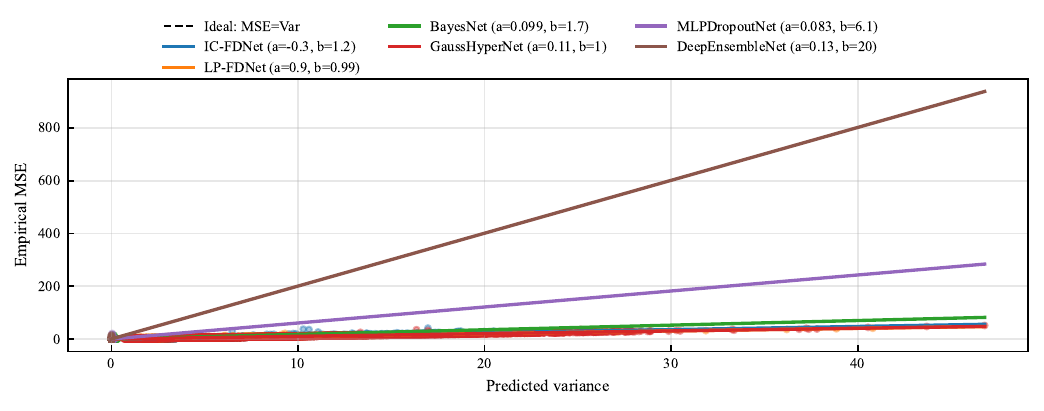}
    \subcaption{CCPP Power Plant}
    \label{fig:real-msevar-ccpp}
  \end{subfigure}

  \vspace{0.4em}

  \begin{subfigure}[t]{0.96\textwidth}
    \centering
    \includegraphics[width=0.95\linewidth]{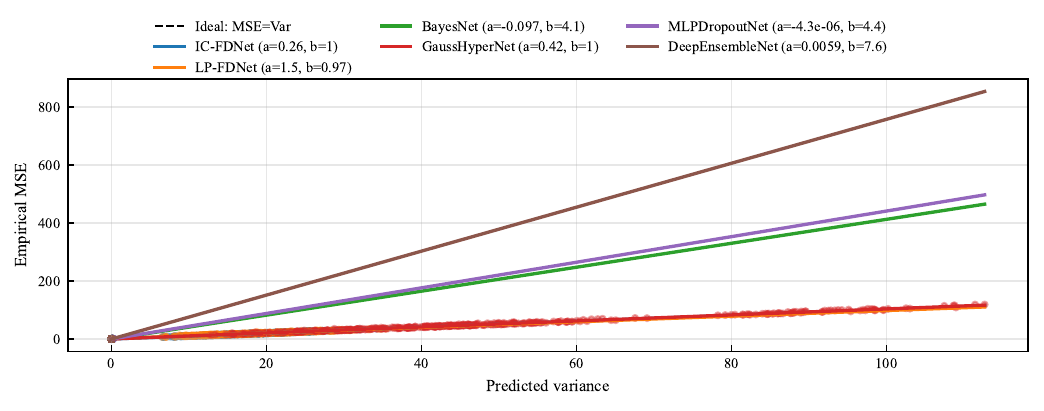}
    \subcaption{Energy Efficiency}
    \label{fig:real-msevar-energy}
  \end{subfigure}

  \caption{MSE vs.\ predicted variance scatter plots for the three real datasets.
  The dashed line marks $\mathrm{MSE}=\mathrm{Var}$; legends report Spearman’s
  $\rho$ and linear-fit parameters.}
  \label{fig:real-msevar-3x1}
\end{figure*}

\begin{figure*}
  \centering
  \captionsetup[subfigure]{justification=centering,singlelinecheck=false,skip=2pt}

  \begin{subfigure}[t]{0.96\textwidth}
    \centering
    \includegraphics[width=0.95\linewidth]{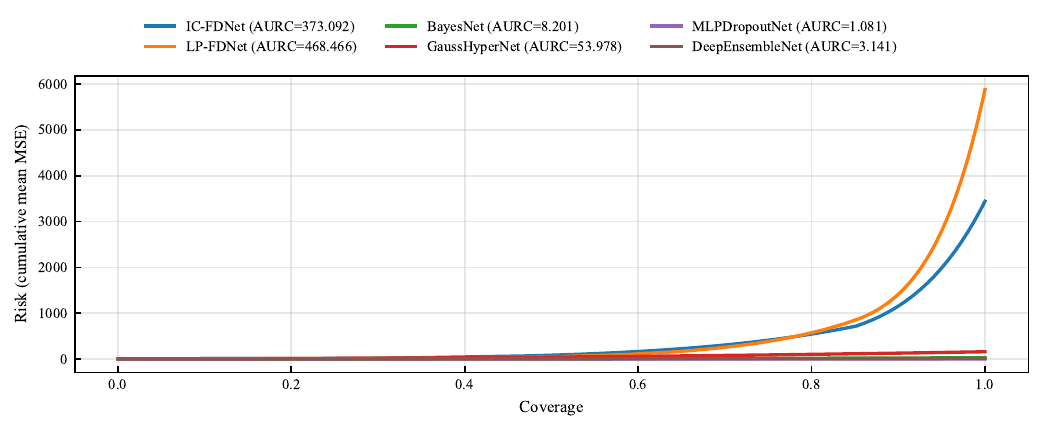}
    \subcaption{Step: $H(x)$}
    \label{fig:toy-aurc-step}
  \end{subfigure}

  \vspace{0.4em}

  \begin{subfigure}[t]{0.96\textwidth}
    \centering
    \includegraphics[width=0.95\linewidth]{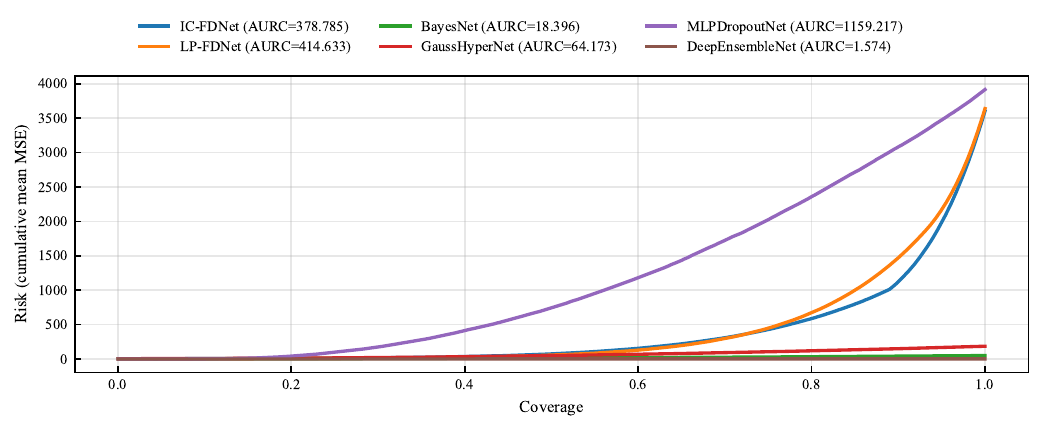}
    \subcaption{Sine: $1.54\,\sin(2.39\,x)$}
    \label{fig:toy-aurc-sine}
  \end{subfigure}

  \vspace{0.4em}

  \begin{subfigure}[t]{0.96\textwidth}
    \centering
    \includegraphics[width=0.95\linewidth]{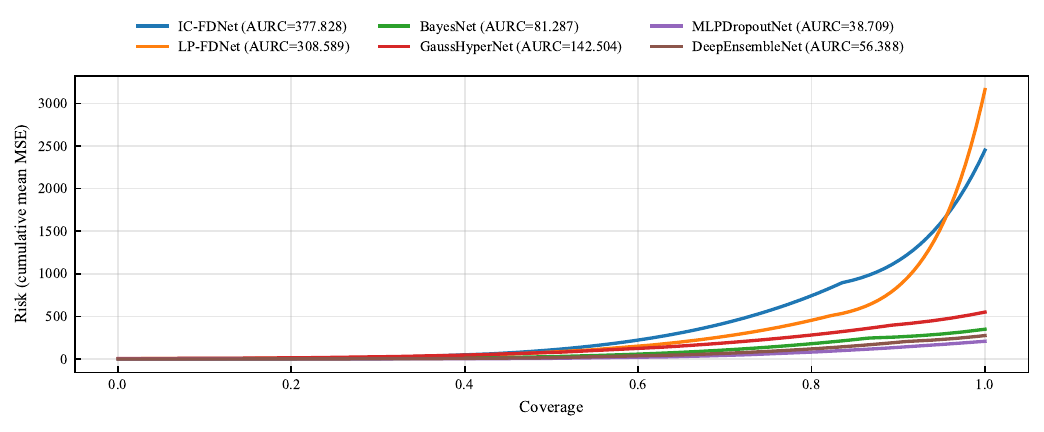}
    \subcaption{Quadratic: $0.43x^{2}-0.41$}
    \label{fig:toy-aurc-quad}
  \end{subfigure}

  \caption{Risk--coverage curves (AURC) for the three toy tasks. Curves plot
  average squared error as a function of coverage as high-variance predictions
  are rejected. Lower AURC indicates better selective regression.}
  \label{fig:toy-aurc-3x1}
\end{figure*}

\begin{figure*}
  \centering
  \captionsetup[subfigure]{justification=centering,singlelinecheck=false,skip=2pt}

  \begin{subfigure}[t]{0.96\textwidth}
    \centering
    \includegraphics[width=0.95\linewidth]{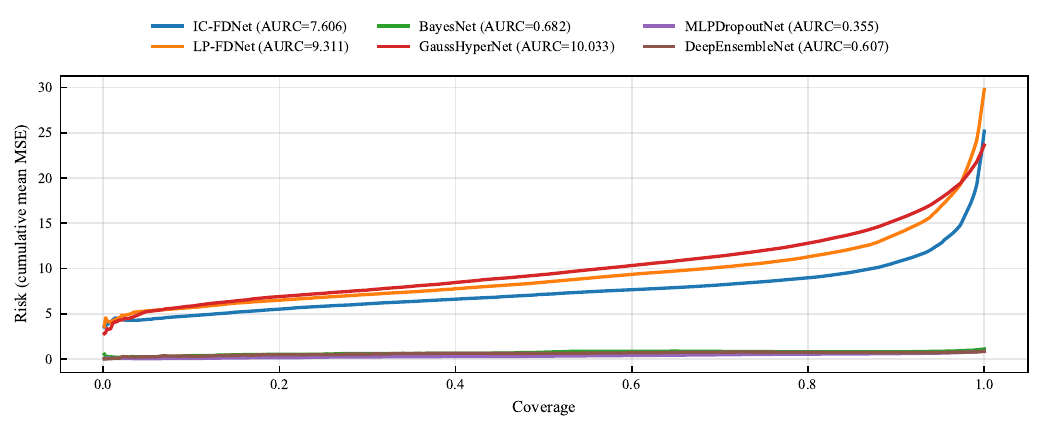}
    \subcaption{Airfoil Self-Noise}
    \label{fig:real-aurc-airfoil}
  \end{subfigure}

  \vspace{0.4em}

  \begin{subfigure}[t]{0.96\textwidth}
    \centering
    \includegraphics[width=0.95\linewidth]{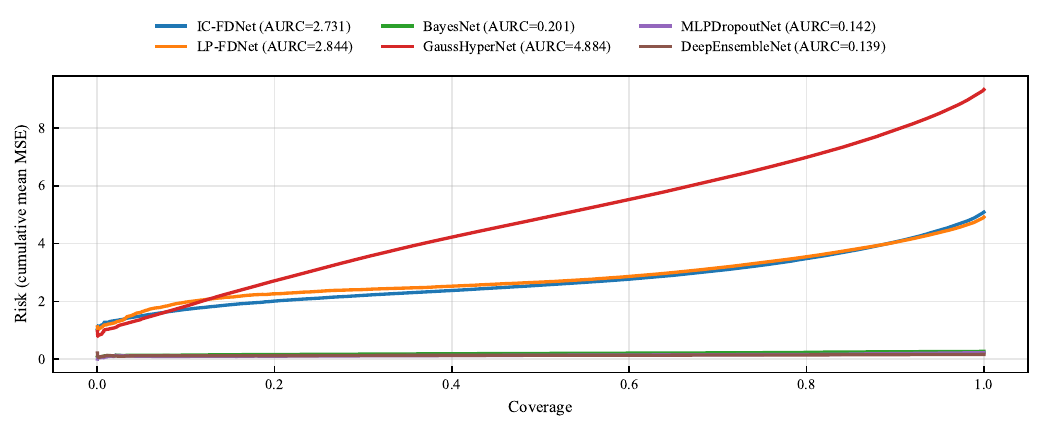}
    \subcaption{CCPP Power Plant}
    \label{fig:real-aurc-ccpp}
  \end{subfigure}

  \vspace{0.4em}

  \begin{subfigure}[t]{0.96\textwidth}
    \centering
    \includegraphics[width=0.95\linewidth]{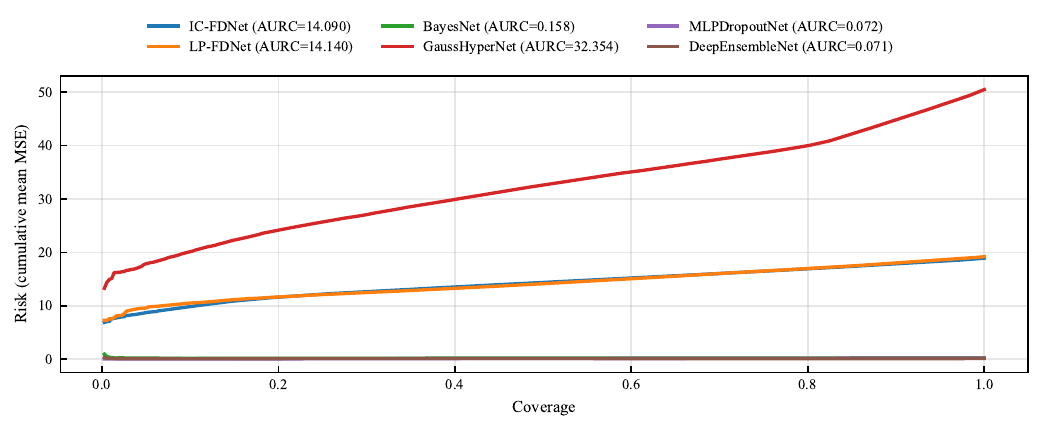}
    \subcaption{Energy Efficiency}
    \label{fig:real-aurc-energy}
  \end{subfigure}

  \caption{Risk--coverage curves (AURC) for the three real datasets. Lower area
  under the curve indicates better ability to abstain on high-error points as
  coverage decreases.}
  \label{fig:real-aurc-3x1}
\end{figure*}

In this appendix we provide additional qualitative diagnostics. 
Figure~\ref{fig:toy-mean-3x1} and Figure~\ref{fig:real-mean-3x1} plot predictive means versus input for the toy and real tasks, respectively. Figure~\ref{fig:toy-msevar-3x1} and Figure~\ref{fig:real-msevar-3x1} show aggregated MSE--variance scatter plots complementing Figure~\ref{fig:toy-msevar-grid} and Figure~\ref{fig:real-msevar-grid} in the main text, while Figure~\ref{fig:toy-aurc-3x1} and Figure~\ref{fig:real-aurc-3x1} report full 
risk--coverage curves (AURC) for toy and real datasets, complementing the summary statistics in Table~\ref{tab:step-unified}, Table~\ref{tab:sine-unified}, Table~\ref{tab:quad-unified}, Table~\ref{tab:airfoil-unified}, Table~\ref{tab:ccpp-unified}, and Table~\ref{tab:energy-unified}.

\FloatBarrier


\begin{table}[H]
\centering
\caption{Monte Carlo and training Hyperparameters (defaults unless noted).}
\label{tab:mc_k}
\begin{tabular}{p{5cm}lc}
\toprule
\textbf{Component} & \textbf{Symbol} & \textbf{Setting} \\
\midrule
MC samples (train)       & $K_{\text{train}}$       & $1$ \\
MC samples (validate)    & $K_{\text{val}}$         & $100$ \\
MC samples (test)        & $K_{\text{test}}$        & $100$ \\
Epochs                   & $\epsilon$               & $400$ \\
Optimizer                & ---                      & ADAM\\
Learning rate            & $\eta$                   & $1 \times 10^{-3}$ \\
Batch size               & $B$                      & $64$ \\
Weight prior std         & $\sigma_0$               & $1.0$ \\
Variance floor           & $\varepsilon$            & $10^{-3}$ in $\sigma = \varepsilon + \mathrm{softplus}(\rho)$ \\
KL schedule              & $\beta_t$                & linear \\
Maximum $\beta$          & $\beta_{\max}$           & $1.0$ \\
Warm up updates           & ---                      & $200$ \\
Toy Grid Seed                    & ---                      & $0$ \\
Step Seed                    & ---                      & $13$ \\
Sine Seed                    & ---                      & $19$ \\
Quadratic Seed                    & ---                      & $6$ \\
Airfoil Self-Noise Seed                    & ---                      & $64$ \\
CCPP Power Plant Seed                    & ---                      & $2$ \\
Energy Efficiency Heating Seed                    & ---                      & $2$ \\
Stochastic Checkpoint          & ---                      & minimum CRPS in interpolation \\
Deterministic Checkpoint           & ---                      & minimum MSE in interpolation\\
\bottomrule
\end{tabular}
\end{table}

\begin{table}[H]
\centering
\caption{Number of training, validation, and test examples used for the toy
1D regression tasks and the three UCI real datasets.}
\label{tab:data_splits}
\begin{tabular}{lrrrr}
\toprule
Dataset & $N_{\text{train}}$ & $N_{\text{val}}$ & $N_{\text{test}}$ & $N_{\text{total}}$ \\
\midrule
Toy 1D functions & 1024 & 512 & 2001 & 3537 \\
Airfoil Self-Noise                    & 597  & 199 & 707  & 1503 \\
CCPP Power Plant                      & 3444 & 1148 & 4976 & 9568 \\
Energy Efficiency (heating load)      & 307  & 102 & 359  & 768  \\
\bottomrule
\end{tabular}
\end{table}

\begin{table}[H]
\centering
\caption{Model configurations and compute. Columns: base hidden width $d_{\text{hid}}$; hypernetwork hidden width $d_{\text{hyper}}$; latent dim $d_h$ (for Gaussian Hypernetwork); ensemble size $M$; parameter count $P$ (per model). Use “—” where not applicable.}
\label{tab:model_grid}
\begingroup
\small
\setlength{\tabcolsep}{6pt}
\begin{tabular}{lcccccccc}
\toprule
\textbf{Model} & $d_{\text{hid}}$ & $d_{\text{hyper}}$ & $d_h$ & $M$ & $P$  \\
\midrule
MLPDropoutNet & \texttt{333} & --- & --- & 1 & \texttt{1000} \\
Deep Ensemble & \texttt{64} & --- & --- & \texttt{10} & \texttt{1000} \\
BayesNet & \texttt{166} & --- & --- & 1 & \texttt{998} \\
Gaussian HyperNet & \texttt{24} & \texttt{5} & \texttt{9} & 1 & \texttt{994}  \\
IC-FDNet & \texttt{23} & \texttt{6} & --- & 1 & \texttt{1004}  \\
LP-FDNet & \texttt{24} & \texttt{5} & --- & 1 & \texttt{1011} \\
\bottomrule
\end{tabular}
\endgroup

\vspace{3pt}
\raggedright
\footnotesize
\textbf{Notes.} 
We can see that the parameter count is roughly equal. In order to keep a fair comparison we scale the number of epochs by the ensemble size so the number of updates is roughly the same.
\end{table}

\begin{table}
\centering
\scriptsize
\setlength{\tabcolsep}{2.5pt}
\renewcommand{\arraystretch}{0.95}
\caption{Step function: unified calibration/uncertainty summary. Lower is better for AURC and deltas ($\Delta{=}$OOD--ID); ideal MSE--Var fit has $a\!\approx\!0$, $b\!\approx\!1$.}
\label{tab:step-unified}
\begin{adjustbox}{max width=\columnwidth}
\begin{tabular}{
l
S[table-format=1.3]   
S[table-format=2.2]   
S[table-format=+3.2]  
S[table-format=3.1]   
S[table-format=4.1]   
S[table-format=4.1]   
S[table-format=2.3]   
}
\toprule
{Model} & {$\rho$} & {$b$} & {$a$} & {AURC $\downarrow$} &
{\makecell{$\Delta$Var\\(OOD--ID) $\uparrow$}} &
{\makecell{$\Delta$MSE\\(OOD--ID) $\downarrow$}} &
{\makecell{$\Delta$CRPS\\(OOD--ID) $\downarrow$}} \\
\midrule
MLPDropoutNet   & 0.990 &  1.34 &   -0.02 &   1.1 &    2.7 &    3.6 &  0.433 \\
DeepEnsembleNet & 0.986 &  4.39 &   -0.08 &   3.1 &    2.6 &   11.3 &  1.579 \\
BayesNet        & 0.987 & 71.82 &   -3.50 &   8.2 &    0.4 &   29.6 &  4.081 \\
GaussHyperNet   & 1.000 &  1.02 &    0.59 &  54.0 &  167.3 &  169.8 &  2.386 \\
IC-FDNet        & 0.992 &  2.52 & -475.80 & 373.1 & 1725.4 & 3820.5 & 16.230 \\
LP-FDNet        & 1.000 &  1.04 &  -17.93 & 468.5 & 6306.7 & 6529.4 &  8.774 \\
\bottomrule
\end{tabular}
\end{adjustbox}
\end{table}

\begin{table}
\centering
\scriptsize
\setlength{\tabcolsep}{2.5pt}
\renewcommand{\arraystretch}{0.95}
\caption{Sine function: unified calibration/uncertainty summary. Lower is better for AURC and deltas ($\Delta{=}$OOD--ID); ideal MSE--Var fit has $a\!\approx\!0$, $b\!\approx\!1$.}
\label{tab:sine-unified}
\begin{adjustbox}{max width=\columnwidth}
\begin{tabular}{
l
S[table-format=1.3]   
S[table-format=2.2]   
S[table-format=+3.2]  
S[table-format=4.1]   
S[table-format=4.1]   
S[table-format=4.1]   
S[table-format=+2.3]  
}
\toprule
{Model} & {$\rho$} & {$b$} & {$a$} & {AURC $\downarrow$} &
{\makecell{$\Delta$Var\\(OOD--ID) $\uparrow$}} &
{\makecell{$\Delta$MSE\\(OOD--ID) $\downarrow$}} &
{\makecell{$\Delta$CRPS\\(OOD--ID) $\downarrow$}} \\
\midrule
MLPDropoutNet   & 0.966 & 32.68 &  476.11 & 1178.4 &   112.6 &  4209.7 &  50.061 \\
DeepEnsembleNet & 0.632 &  1.16 &    1.16 &    1.5 &     1.0 &     1.1 &  -0.179 \\
BayesNet        & 0.999 &  1.00 &    1.20 &   19.7 &    55.7 &    55.7 &   1.066 \\
GaussHyperNet   & 1.000 &  1.02 &    0.85 &   59.0 &   183.8 &   187.0 &   2.265 \\
IC-FDNet        & 0.973 &  1.73 & 1098.34 &  623.4 &  1436.3 &  3705.1 &  20.842 \\
LP-FDNet        & 0.999 &  1.05 &   38.14 &  382.9 &  3170.0 &  3362.4 &   6.527 \\
\bottomrule
\end{tabular}
\end{adjustbox}
\end{table}

\begin{table}
\centering
\scriptsize
\setlength{\tabcolsep}{2.5pt}
\renewcommand{\arraystretch}{0.95}
\caption{Quadratic function: unified calibration/uncertainty summary. Lower is better for AURC and deltas ($\Delta{=}$OOD--ID); ideal MSE--Var fit has $a\!\approx\!0$, $b\!\approx\!1$.}
\label{tab:quad-unified}
\begin{adjustbox}{max width=\columnwidth}
\begin{tabular}{
l
S[table-format=1.3]   
S[table-format=4.2]   
S[table-format=+3.2]  
S[table-format=4.1]   
S[table-format=4.1]   
S[table-format=4.1]   
S[table-format=2.3]   
}
\toprule
{Model} & {$\rho$} & {$b$} & {$a$} & {AURC $\downarrow$} &
{\makecell{$\Delta$Var\\(OOD--ID) $\uparrow$}} &
{\makecell{$\Delta$MSE\\(OOD--ID) $\downarrow$}} &
{\makecell{$\Delta$CRPS\\(OOD--ID) $\downarrow$}} \\
\midrule
MLPDropoutNet   & 0.990 & 3459.12 &  -67.32 &  38.7 &    0.1 &  231.7 & 11.176 \\
DeepEnsembleNet & 0.979 &  811.63 &  -68.32 &  56.4 &    0.5 &  306.2 & 13.003 \\
BayesNet        & 0.953 & 3855.46 &  -89.49 &  81.3 &    0.1 &  389.1 & 15.136 \\
GaussHyperNet   & 0.993 &    2.81 & -112.41 & 142.5 &  253.9 &  603.6 &  9.137 \\
IC-FDNet        & 0.988 &    1.42 &  225.67 & 377.8 & 1743.9 & 2721.7 & 14.073 \\
LP-FDNet        & 0.997 &    1.42 & -107.08 & 308.6 & 2547.3 & 3510.6 & 10.325 \\
\bottomrule
\end{tabular}
\end{adjustbox}
\end{table}

\begin{table}
\centering
\scriptsize
\setlength{\tabcolsep}{2.5pt}
\renewcommand{\arraystretch}{0.95}
\caption{Airfoil Self-Noise: unified calibration/uncertainty summary. Lower is better for AURC and deltas ($\Delta{=}$OOD--ID); ideal MSE--Var fit has $a\!\approx\!0$, $b\!\approx\!1$.}
\label{tab:airfoil-unified}
\begin{adjustbox}{max width=\columnwidth}
\begin{tabular}{
l
S[table-format=1.3]   
S[table-format=2.2]   
S[table-format=+3.2]  
S[table-format=3.1]   
S[table-format=4.1]   
S[table-format=4.1]   
S[table-format=2.3]   
}
\toprule
{Model} & {$\rho$} & {$b$} & {$a$} & {AURC $\downarrow$} &
{\makecell{$\Delta$Var\\(OOD--ID) $\uparrow$}} &
{\makecell{$\Delta$MSE\\(OOD--ID) $\downarrow$}} &
{\makecell{$\Delta$CRPS\\(OOD--ID) $\downarrow$}} \\
\midrule
MLPDropoutNet   & 0.601 & 10.14 &  0.21 &  0.4 &   0.1 &   1.1 & 0.535 \\
DeepEnsembleNet & 0.177 &  9.74 &  0.57 &  0.6 &   0.0 &   0.8 & 0.353 \\
BayesNet        & 0.330 &  4.49 &  0.40 &  0.7 &   0.2 &   1.1 & 0.335 \\
GaussHyperNet   & 0.987 &  1.01 &  0.90 & 10.0 &  18.9 &  19.1 & 0.340 \\
IC-FDNet        & 0.976 &  1.40 & -3.40 &  7.6 &  19.4 &  25.3 & 0.493 \\
LP-FDNet        & 0.983 &  1.00 &  1.01 &  9.3 &  28.5 &  28.8 & 0.363 \\
\bottomrule
\end{tabular}
\end{adjustbox}
\end{table}

\begin{table}
\centering
\scriptsize
\setlength{\tabcolsep}{2.5pt}
\renewcommand{\arraystretch}{0.95}
\caption{CCPP Power Plant: unified calibration/uncertainty summary. Lower is better for AURC and deltas ($\Delta{=}$OOD--ID); ideal MSE--Var fit has $a\!\approx\!0$, $b\!\approx\!1$.}
\label{tab:ccpp-unified}
\begin{adjustbox}{max width=\columnwidth}
\begin{tabular}{
l
S[table-format=1.3]   
S[table-format=2.2]   
S[table-format=+1.3]  
S[table-format=1.3]   
S[table-format=1.3]   
S[table-format=1.3]   
S[table-format=1.3]   
}
\toprule
{Model} & {$\rho$} & {$b$} & {$a$} & {AURC $\downarrow$} &
{\makecell{$\Delta$Var\\(OOD--ID) $\uparrow$}} &
{\makecell{$\Delta$MSE\\(OOD--ID) $\downarrow$}} &
{\makecell{$\Delta$CRPS\\(OOD--ID) $\downarrow$}} \\
\midrule
MLPDropoutNet   & 0.417 &  6.07 &  0.083 & 0.142 & 0.023 & 0.156 & 0.095 \\
DeepEnsembleNet & 0.126 & 20.05 &  0.127 & 0.139 & 0.002 & 0.071 & 0.065 \\
BayesNet        & 0.470 &  1.75 &  0.099 & 0.201 & 0.065 & 0.132 & 0.045 \\
GaussHyperNet   & 0.998 &  1.02 &  0.108 & 4.884 & 7.134 & 7.159 & 0.283 \\
IC-FDNet        & 0.970 &  1.20 & -0.303 & 2.731 & 2.916 & 3.401 & 0.211 \\
LP-FDNet        & 0.902 &  0.99 &  0.903 & 2.844 & 2.129 & 2.820 & 0.260 \\
\bottomrule
\end{tabular}
\end{adjustbox}
\end{table}

\begin{table}
\centering
\scriptsize
\setlength{\tabcolsep}{2.5pt}
\renewcommand{\arraystretch}{0.95}
\caption{Energy Efficiency: unified calibration/uncertainty summary. Lower is better for AURC and deltas ($\Delta{=}$OOD--ID); ideal MSE--Var fit has $a\!\approx\!0$, $b\!\approx\!1$.}
\label{tab:energy-unified}
\begin{adjustbox}{max width=\columnwidth}
\begin{tabular}{
l
S[table-format=1.3]   
S[table-format=2.2]   
S[table-format=+3.2]  
S[table-format=3.1]   
S[table-format=4.1]   
S[table-format=4.1]   
S[table-format=1.3]   
}
\toprule
{Model} & {$\rho$} & {$b$} & {$a$} & {AURC $\downarrow$} &
{\makecell{$\Delta$Var\\(OOD--ID) $\uparrow$}} &
{\makecell{$\Delta$MSE\\(OOD--ID) $\downarrow$}} &
{\makecell{$\Delta$CRPS\\(OOD--ID) $\downarrow$}} \\
\midrule
MLPDropoutNet   & 0.696 &  4.42 &  -0.00 &  0.1 &   0.0 &   0.3 & 0.258 \\
DeepEnsembleNet & 0.426 &  7.58 &   0.01 &  0.1 &   0.0 &   0.0 & 0.051 \\
BayesNet        & 0.549 &  4.13 &  -0.10 &  0.2 &   0.0 &   0.1 & 0.106 \\
GaussHyperNet   & 0.996 &  1.04 &   0.42 & 32.4 &  30.3 &  30.9 & 0.454 \\
IC-FDNet        & 0.993 &  1.02 &   0.26 & 14.1 &   6.6 &   6.8 & 0.114 \\
LP-FDNet        & 0.989 &  0.97 &   1.45 & 14.1 &   6.8 &   6.2 & 0.005 \\
\bottomrule
\end{tabular}
\end{adjustbox}
\end{table}


\begin{figure}[H]
    \centering
    \includegraphics[width=0.9\linewidth]{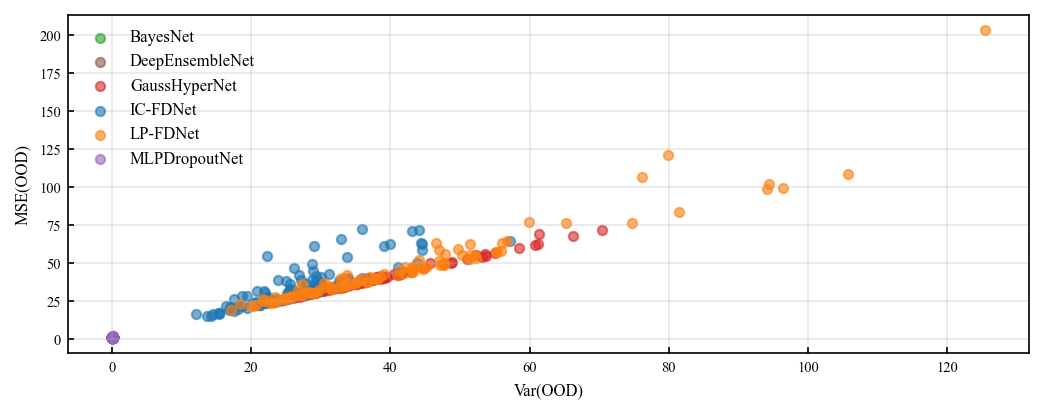}
    \caption{
    Seed-aggregated MSE--variance scatter on the Airfoil Self-Noise dataset over
    100 random initializations. Each point summarizes one seed by its OOD MSE and
    OOD predictive variance, and the representative seed used in the main-text
    plots lies close to the overall across-seed trend.
    }
    \label{fig:seed-agg-mse-ood-var-ood}
\end{figure}

\clearpage       

\section{Computational complexity and scalability}
\label{app:complexity}

\paragraph{Deterministic MLP.}
Let $N$ denote the number of training examples, $d_x$ the input dimension,
$d_y$ the output dimension, and consider a base network with widths
$\{d_\ell\}_{\ell=0}^L$ with $d_0 = d_x$, $d_L = d_y$. A deterministic MLP
has per-epoch cost
\[
O\!\left(
  N \sum_{\ell=1}^L d_{\ell-1} d_\ell
\right),
\]
i.e., linear in $N$ and quadratic in the layer widths.

\paragraph{FDN and Hypernetworks.}
FDN replaces each fixed weight matrix by an input-conditioned diagonal
Gaussian $q_\phi(W_\ell, b_\ell \mid s_\ell(x))$ whose parameters are
generated by a small per-layer Hypernetwork. For layer $\ell$, the
Hypernetwork takes a conditioning vector
\[
s_\ell(x) =
\begin{cases}
x, & \text{IC-FDN},\\
a_{\ell-1}(x), & \text{LP-FDN},
\end{cases}
\]
with dimension $d_{s,\ell} \in \{d_x, d_{\ell-1}\}$, and outputs
$(\mu_{W_\ell}, \log\sigma^2_{W_\ell}, \mu_{b_\ell}, \log\sigma^2_{b_\ell})
\in \mathbb{R}^{P_\ell}$, where $P_\ell = 2(d_{\ell-1} d_\ell + d_\ell)$.
In our implementation, each Hypernetwork $A_\ell$ is a two-layer MLP with
hidden width $h_{\mathrm{hyp}}$. For a mini-batch of size $B$ and $K$ Monte
Carlo samples per example, the per-epoch complexity is
\[
O\!\left(
  N K \sum_{\ell=1}^L
  \bigl(
    d_{s,\ell} h_{\mathrm{hyp}}
    + h_{\mathrm{hyp}} P_\ell
    + d_{\ell-1} d_\ell
  \bigr)
\right).
\]
The three bracketed terms correspond to (i) the Hypernetwork input transform,
(ii) projection to the $P_\ell$ Gaussian parameters, and (iii) the
base-layer matrix–vector product. In all experiments we use $K = 1$ and
small Hypernetwork widths (chosen to satisfy a global $\approx 10^3$
parameter budget), making $h_{\mathrm{hyp}} P_\ell$ comparable to
$d_{\ell-1} d_\ell$. As a result:
(i) the training cost of IC-/LP-FDN remains linear in $N$; 
(ii) the runtime overhead relative to a deterministic MLP is a small
constant factor (typically $2$–$4\times$), set by $h_{\mathrm{hyp}}$ and $K$.

\paragraph{Parameter-count clarification.}
A concern raised in the reviews was that LP-FDN parameter count might scale cubically
in model size (e.g., as $O(d_\ell^3)$ in the width of a layer). This
would occur only for a specific design where the Hypernetwork directly
maps a $d_\ell$-dimensional activation to all $d_\ell^2$ entries of a
dense weight matrix via a fully connected layer, which would require a
$d_\ell^2 \times d_\ell$ matrix.

In our implementation, however, the Hypernetwork hidden width
$h_{\mathrm{hyp}}$ is a small constant, independent of $d_\ell$:
\[
P^{\text{hyper}}_\ell
  = d_{s,\ell} h_{\mathrm{hyp}}
    + h_{\mathrm{hyp}} \cdot 2(d_{\ell-1} d_\ell + d_\ell)
  \sim O\!\bigl(h_{\mathrm{hyp}} d_{\ell-1} d_\ell\bigr)
  \sim O(d_{\ell-1} d_\ell).
\]
Thus IC-FDN and LP-FDN both have the same $O(d_{\ell-1} d_\ell)$
scaling as the corresponding base MLP layer; in particular, there is
no cubic dependence on layer width.

\paragraph{Architectural trade-offs for larger scales.}
FDN is compatible with standard complexity-reduction techniques without
changing the formulation: (i) low-rank factorizations $W_\ell = U_\ell
V_\ell^\top$ with the Hypernetwork generating only $(U_\ell, V_\ell)$;
(ii) row- or column-wise generation instead of full matrices; and
(iii) a shared Hypernetwork with layer-specific output heads. In the main
experiments we keep the architecture minimal to match parameter and update
budgets across baselines, but these options make FDN readily extendable to
higher-dimensional and large-$N$ settings.

\paragraph{Middle-layer usage and adapter-style deployment.}
In practice, Hypernetworks need not be applied to \emph{every} layer of a deep backbone. FDN is
layer-local, so one can restrict stochastic, input-conditioned weights to a small subset of higher
layers (or even just the predictive head), keeping earlier blocks deterministic and frozen. This is
analogous in spirit to LoRA and adapter modules~\citep{hu2022lora,houlsby2019parameter}: a narrow,
trainable “uncertainty adapter’’ is inserted on top of a largely fixed backbone, so the additional
parameters and compute scale with the adapter width rather than with the full network depth or
width. In such configurations the overall parameter and FLOP overhead of FDN remains a small
fraction of the backbone, even for large architectures, while still enabling input-dependent
uncertainty where it is most needed.

\section{Evaluation metrics and calibration diagnostics}
\label{app:metrics-calib}

For a test set $\{(x_i, y_i)\}_{i=1}^T$ and a stochastic predictor that
yields $K$ samples $\{y_i^{(k)}\}_{k=1}^K$ from the predictive
distribution $p(y \mid x_i, \mathcal{D})$, we use the following metrics.

\paragraph{Point prediction error.}
The predictive mean at $x_i$ is
\[
\hat{\mu}_i = \frac{1}{K}\sum_{k=1}^K y_i^{(k)}.
\]
We define the \emph{per-point} Monte Carlo MSE and bias as
\[
\mathrm{MSE}_i = \frac{1}{K} \sum_{k=1}^K \bigl(y_i^{(k)} - y_i\bigr)^2,
\qquad
\mathrm{Bias}_i = \hat{\mu}_i - y_i
= \frac{1}{K} \sum_{k=1}^K \bigl(y_i^{(k)} - y_i\bigr).
\]
For any subset of test inputs $S \subseteq \{1,\dots,T\}$ (e.g., all, ID, or OOD),
we aggregate by averaging over $i \in S$:
\[
\mathrm{MSE}_S = \frac{1}{|S|} \sum_{i \in S} \mathrm{MSE}_i,
\qquad
\mathrm{Bias}_S = \frac{1}{|S|} \sum_{i \in S} \mathrm{Bias}_i.
\]
Unless otherwise noted, reported MSE and Bias refer to these region-averaged
quantities.

\paragraph{Predictive variance and epistemic uncertainty.}
The Monte Carlo estimator of predictive variance at $x_i$ is
\[
\widehat{\mathrm{Var}}[Y \mid x_i]
= \frac{1}{K}\sum_{k=1}^K \bigl(y_i^{(k)} - \hat{\mu}_i\bigr)^2,
\]
which we aggregate over ID or OOD test splits by averaging across $i$.
In the homoscedastic Gaussian setting this decomposes into aleatoric and
epistemic components via the law of total variance; we focus on the
epistemic part induced by the weight distribution.


\paragraph{Continuous ranked probability score (CRPS).}
For a univariate predictive CDF $F_i(y)$ and realization $y_i$, the CRPS is
\[
\mathrm{CRPS}(F_i, y_i)
= \int_{-\infty}^{\infty} \!\bigl(F_i(z) - \mathbf{1}\{z \ge y_i\}\bigr)^2 dz.
\]
Using samples $y_i^{(k)} \sim F_i$, we apply the standard Monte Carlo
estimator
\[
\widehat{\mathrm{CRPS}}_i
= \frac{1}{K}\sum_{k=1}^K |y_i^{(k)} - y_i|
  - \frac{1}{2K^2}\sum_{k=1}^K \sum_{\ell=1}^K
    |y_i^{(k)} - y_i^{(\ell)}|.
\]
We report the average CRPS over ID and OOD test splits.
Lower CRPS within a region (ID or OOD) indicates a sharper and better
calibrated predictive distribution: mass is concentrated near $y_i$
without being spuriously overconfident.
Under shift we typically expect $\mathrm{CRPS}_{\text{OOD}} >
\mathrm{CRPS}_{\text{ID}}$ because the task is harder; for a fixed OOD
difficulty, smaller $\mathrm{CRPS}_{\text{OOD}}$ (or smaller
$\Delta\mathrm{CRPS}$, see below) is better.

\paragraph{Calibration: MSE--variance relation.}
To assess calibration we compare the predicted variance
$\widehat{\mathrm{Var}}[Y \mid x_i]$ with the empirical per-point MSE
\[
e_i = \mathrm{MSE}_i
= \frac{1}{K} \sum_{k=1}^K \bigl(y_i^{(k)} - y_i\bigr)^2.
\]
We first compute Spearman rank correlation
$\rho = \mathrm{corr}_{\mathrm{Spearman}}(\{\widehat{\mathrm{Var}}_i\},
\{e_i\})$ and fit the linear relation
$e_i \approx a + b\,\widehat{\mathrm{Var}}_i$ by least squares; the ideal
fit has $a \approx 0$ and $b \approx 1$.
High $\rho$ means that larger predicted variance reliably flags larger
squared error (good ranking), while $(a,b)$ measure the \emph{scale}
of the variances relative to the errors.

In addition, we form \emph{variance--MSE calibration curves} by binning
test points into $B$ quantiles of predicted variance. In each bin $b$
(with index set $S_b$) we compute the mean predicted variance
$\mathrm{Var}_b = \frac{1}{|S_b|}\sum_{i \in S_b} \widehat{\mathrm{Var}}_i$
and empirical MSE
$\mathrm{MSE}_b = \frac{1}{|S_b|}\sum_{i \in S_b} \mathrm{MSE}_i$,
and plot the pairs $(\mathrm{Var}_b,\mathrm{MSE}_b)$ together with the
ideal $y{=}x$ line. Points lying close to this diagonal indicate that the
typical error magnitude in each confidence bin matches the predicted
variance scale.

\paragraph{Risk--coverage (AURC).}
Using variance as an inverse-confidence score, we sort test points by
increasing $\widehat{\mathrm{Var}}[Y \mid x_i]$. For a coverage level
$c \in (0,1]$ (fraction of most-confident points retained) we compute the
cumulative risk $R(c)$ as the average per-point MSE over the retained subset:
\[
R(c) = \frac{1}{|S(c)|} \sum_{i \in S(c)} \mathrm{MSE}_i,
\]
where $S(c)$ contains the most-confident fraction $c$ of test points.
The area under the risk--coverage curve,
$\mathrm{AURC} = \int_0^1 R(c)\, dc$, is estimated numerically.
Lower AURC is better: for a fixed difficulty, it means that as we keep
only high-confidence predictions, the resulting risk drops more quickly.

\paragraph{ID vs.\ OOD deltas.}
For each model and dataset we compute MSE, variance, and CRPS separately
on ID (interpolation) and OOD (extrapolation) regions, using the region
averages defined above, and report
\[
\Delta\mathrm{MSE}
= \mathrm{MSE}_{\text{OOD}} - \mathrm{MSE}_{\text{ID}},\quad
\Delta\mathrm{Var}
= \mathrm{Var}_{\text{OOD}} - \mathrm{Var}_{\text{ID}},\quad
\Delta\mathrm{CRPS}
= \mathrm{CRPS}_{\text{OOD}} - \mathrm{CRPS}_{\text{ID}}.
\]
For a fixed notion of shift, good uncertainty estimates should exhibit
small $\Delta\mathrm{MSE}$ (robust accuracy), \emph{large and positive}
$\Delta\mathrm{Var}$ (higher uncertainty OOD than ID), and small
$\Delta\mathrm{CRPS}$ (predictive distributions that degrade gracefully
rather than collapsing or becoming wildly miscalibrated).

\paragraph{On NLL / NLPD.}
Negative log predictive density (NLL / NLPD) is another strictly proper
scoring rule for probabilistic regression and is closely related to CRPS.
We computed NLL in preliminary experiments, but found that in our setting
it was (i) strongly correlated with CRPS and MSE, and (ii) much more
sensitive to occasional extreme errors due to the logarithm, which can
dominate the average and obscure more typical behavior. CRPS, by contrast,
remains finite, can be estimated directly from samples without specifying
a parametric density or bandwidth, and provides a more interpretable
summary of the overall predictive distribution (both sharpness and
calibration) under dataset shift. For these reasons, and to avoid redundant
plots/tables, we report CRPS (together with MSE, variance, and AURC) as our
primary proper scoring rule and omit NLL/NLPD from the main results.


\end{document}